\documentclass[pdflatex,sn-mathphys-num]{sn-jnl}

\usepackage{graphicx}
\usepackage{multirow}
\usepackage{amsmath,amssymb,amsfonts}
\usepackage{amsthm}
\usepackage{mathrsfs}
\usepackage[title]{appendix}
\usepackage{textcomp}
\usepackage{manyfoot}
\usepackage{booktabs}
\usepackage{algorithm}
\usepackage{algorithmicx}
\usepackage{algpseudocode}
\usepackage{listings}
\usepackage{siunitx}
\usepackage{comment}
\usepackage{caption}
\usepackage{subcaption}
\usepackage{hyperref}
\usepackage[table,xcdraw]{xcolor}

\usepackage[utf8]{inputenc}
\usepackage[british]{babel}
\usepackage{geometry}
\usepackage{natbib}
\usepackage{array}
\usepackage{pifont}
\usepackage{makecell}
\usepackage{tabularx}
\usepackage{mathtools}

\hypersetup{
    colorlinks=true,
    linkcolor=blue,
    filecolor=magenta,
    urlcolor=cyan,
    citecolor=blue,
}

\theoremstyle{thmstyleone}

\theoremstyle{thmstyletwo}

\theoremstyle{thmstylethree}

\begin{document}

\title[Article Title]{SurvBench: A Standardised Preprocessing Pipeline for Multi-Modal Electronic Health Record Survival Analysis}

\author*[1]{\fnm{Munib} \sur{Mesinovic}}\email{munib.mesinovic@eng.ox.ac.uk}

\author[1]{\fnm{Tingting} \sur{Zhu}}

\affil*[1]{\orgdiv{Department of Engineering Science}, \orgname{University of Oxford}, \orgaddress{\city{Oxford}, \country{UK}}}

\abstract{
\noindent\textbf{BACKGROUND}\quad Deep-learning survival models for electronic health record (EHR) data are hard to compare across papers because the upstream preprocessing step, which includes cohort definition, time discretisation, missingness handling, and censoring rules, is typically undocumented and inconsistent. A reported difference in concordance between two mortality models can therefore reflect any of these choices rather than a modelling contribution.

\noindent\textbf{METHODS}\quad We present SurvBench, an open-source preprocessing pipeline that converts raw PhysioNet exports into model-ready tensors for survival analysis. SurvBench covers four critical-care databases (MIMIC-IV, eICU, MC-MED, HiRID) and four input modalities: time-series vitals and laboratory values, static demographics, International Classification of Diseases (ICD) codes, and radiology report embeddings. Every preprocessing decision is controlled through YAML configuration. Imputation, scaling, and feature filtering are fit on the training fold only. Missingness is recorded as a binary mask alongside each feature tensor. The pipeline handles single-risk endpoints (in-hospital and in-ICU mortality) and competing-risks endpoints (a three-way emergency-department admission pathway, with home discharge treated as administrative censoring). We also provide support for harmonised cross-dataset external validation between eICU and MIMIC-IV.

\noindent\textbf{RESULTS}\quad SurvBench produces cohorts ranging from 15{,}343 ICU stays in HiRID to 127{,}977 in eICU, with training event rates between 4.3\% and 7.5\% on the three single-risk ICU mortality tasks and 40.6\% on the MC-MED admission-pathway task. Outputs are compatible with pycox and standard PyTorch survival code. To use every modality the pipeline produces, we trained five reference baselines (Cox proportional hazards, DeepHit, Dynamic-DeepHit, DySurv, and a new multi-modal transformer, TransformerSurv) and evaluated them under Antolini's time-dependent concordance, integrated Brier score, integrated negative binomial log-likelihood, and cumulative dynamic AUC. Across five seeds, Dynamic-DeepHit achieved the highest test-set concordance on MIMIC-IV and HiRID ($C^{\mathrm{td}}$ of 0.842 and 0.862), TransformerSurv (time-series and static) achieved the highest concordance on eICU ($C^{\mathrm{td}} = 0.796$), and the multi-modal TransformerSurv achieved the highest concordance on the MC-MED competing-risks task ($C^{\mathrm{td}} = 0.774$). On cross-dataset transfer between eICU and MIMIC-IV, the Transformer achieves the highest out-of-distribution discrimination (C$^{td}$ = 0.812 $\pm$ 0.007, AUC$_{int}$ = 0.732 $\pm$ 0.009).

\noindent\textbf{CONCLUSIONS}\quad SurvBench is publicly available at \url{https://github.com/munibmesinovic/SurvBench}, providing a robust platform that future deep-learning EHR survival work, especially nascent multi-modal approaches, can be measured against under matched preprocessing.
}


\maketitle


\section*{Introduction}

Survival analysis models the time from a baseline event, such as admission or diagnosis, to an outcome such as death, discharge, or recurrence, while handling patients whose outcome has not yet been observed \citep{ranganath2016deep}. Deep-learning extensions of the classical proportional-hazards and discrete-time hazard frameworks have been proposed to capture the non-linear temporal patterns common in electronic health record (EHR) data \citep{wiegrebe2024deep}. Progress has been held back by what we call the preprocessing gap. The code that converts raw EHR comma-separated value (CSV) exports into model inputs is rarely shared, rarely documented in full, and varies between papers in ways that could affect reported performance. Two papers reporting concordance indices on ``MIMIC-IV mortality'' may compute them on different cohorts, with different censoring rules, different temporal aggregations, and different imputation strategies \citep{misra2019impact}. A difference in performance can therefore reflect any of these choices rather than a modelling contribution.

EHRs pose specific challenges for survival analysis, and each preprocessing decision has a measurable effect on downstream performance \citep{misra2019impact}. Temporal aggregation can be hourly, fixed-width, or event-driven, and each strategy trades temporal resolution against computational cost \citep{manojlovic2017efficient}. Missingness can be imputed, masked, or both \citep{ren2024moving}. Feature selection requires a prevalence threshold below which sparsely measured variables are dropped, trading dimensionality against keeping informative-but-rare measurements \citep{wang2024mel}. Normalisation has to bring features onto comparable ranges without flattening clinically meaningful variation \citep{singh2022feature}. The outcome definition and censoring rule have to align with a clinical endpoint while accommodating administrative censoring and competing risks \citep{gregson2024competing, thomas2021competing}. Splitting into training, validation, and test has to operate at the patient level so that repeated encounters by the same individual do not leak across folds.

The deep-learning survival literature has expanded rapidly in recent years, with parametric models (DeepSurv) \citep{katzman2018deepsurv}, competing-risks frameworks (DeepCompete) \citep{huang2021deepcompete}, recurrence-based architectures (Dynamic-DeepHit) \citep{lee2019dynamic}, and conditional variational autoencoders for dynamic risk prediction (DySurv) \citep{mesinovic2026dysurv}. Each typically introduces a custom preprocessing pipeline tailored to a specific dataset and task, with limited documentation of the implementation. Reproducing a baseline often requires reverse-engineering an undocumented preparation step from an incomplete description.

The widely cited MIMIC-III benchmark by \citep{wang2020mimic} provides preprocessed data for multiple prediction tasks, but is built on a database that has now been superseded by MIMIC-IV and does not support survival analysis or competing risks. Recent work has explored MIMIC-IV \citep{nguyen2023mimic} for various prediction tasks \citep{lovon2024revisiting, bui2024benchmarking}, but standardised survival-analysis pipelines are absent. The eICU Collaborative Research Database \citep{pollard2018eicu} has received less attention despite covering more than 200{,}000 ICU admissions across 335 hospitals. To our knowledge, no widely adopted public benchmark for time-to-event survival analysis, including competing risks, exists for MIMIC-IV, eICU, MC-MED, or HiRID, especially providing robust multi-modality compatibility.

SurvBench is primarily a preprocessing pipeline. While we provide saved model checkpoints, the goal is not to ship pretrained models, leaderboard tables, or fixed evaluation splits intended for inter-paper comparison. We provide a reproducible mapping from raw PhysioNet CSVs to model-ready tensors, governed end-to-end by customisable YAML configuration, with patient-level stratified splits, explicit missingness masks, training-fold-only fitting of imputers and scalers, and identical output schemas across the supported databases. We additionally include a small set of trained reference baselines, including a new multi-modal transformer (TransformerSurv), that exercise every modality the pipeline produces. These are sanity checks and starting points for new work. The shipped configurations target related but distinct clinical questions across ICU and emergency department (ED) settings. ICU mortality on MIMIC-IV, eICU, and HiRID is a single-risk endpoint with a 240-hour horizon and training event rates between 4\% and 8\%. ED disposition on MC-MED is a three-way competing-risks endpoint (hospital admission, observation, ICU admission) over a 24-hour horizon, with patients discharged home administratively censored at the time of discharge, giving a training event rate (any admission) of 40.6\%. These are not interchangeable tasks, and results on one are not an entry against the other. SurvBench standardises preprocessing within each task so that comparisons within a task are well defined.

SurvBench is publicly available on GitHub\footnote{\url{https://github.com/munibmesinovic/SurvBench}} under an open-source licence, with documentation, configuration examples, and visualisation scripts.


\section*{SurvBench}

\begin{figure*}[!ht]
\centering
\begin{subfigure}[t]{0.48\textwidth}
\centering
\includegraphics[width=\textwidth]{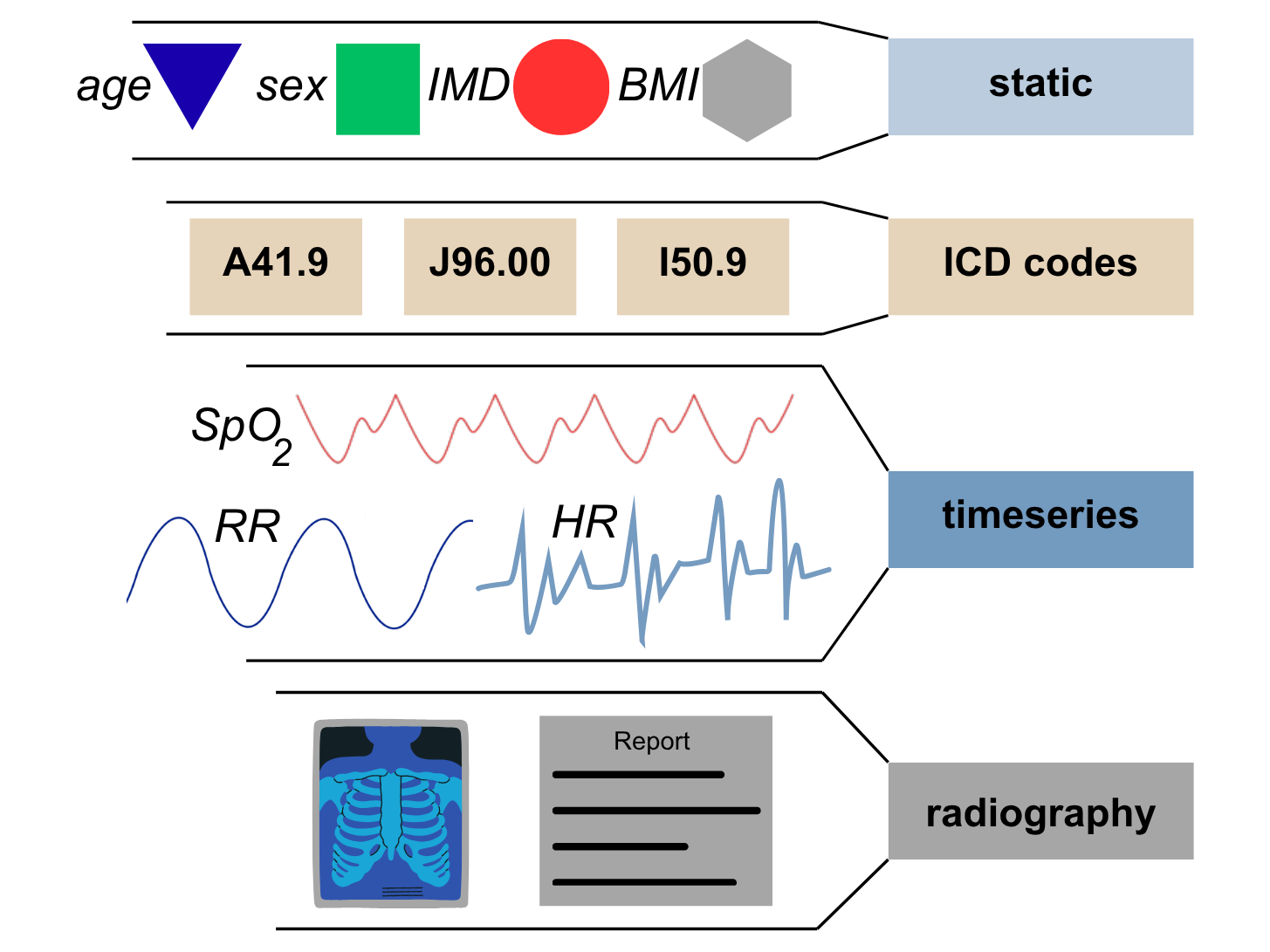}
\caption{}
\label{fig:pipeline_a_modalities}
\end{subfigure}
\hfill
\begin{subfigure}[t]{0.48\textwidth}
\centering
\includegraphics[width=\textwidth]{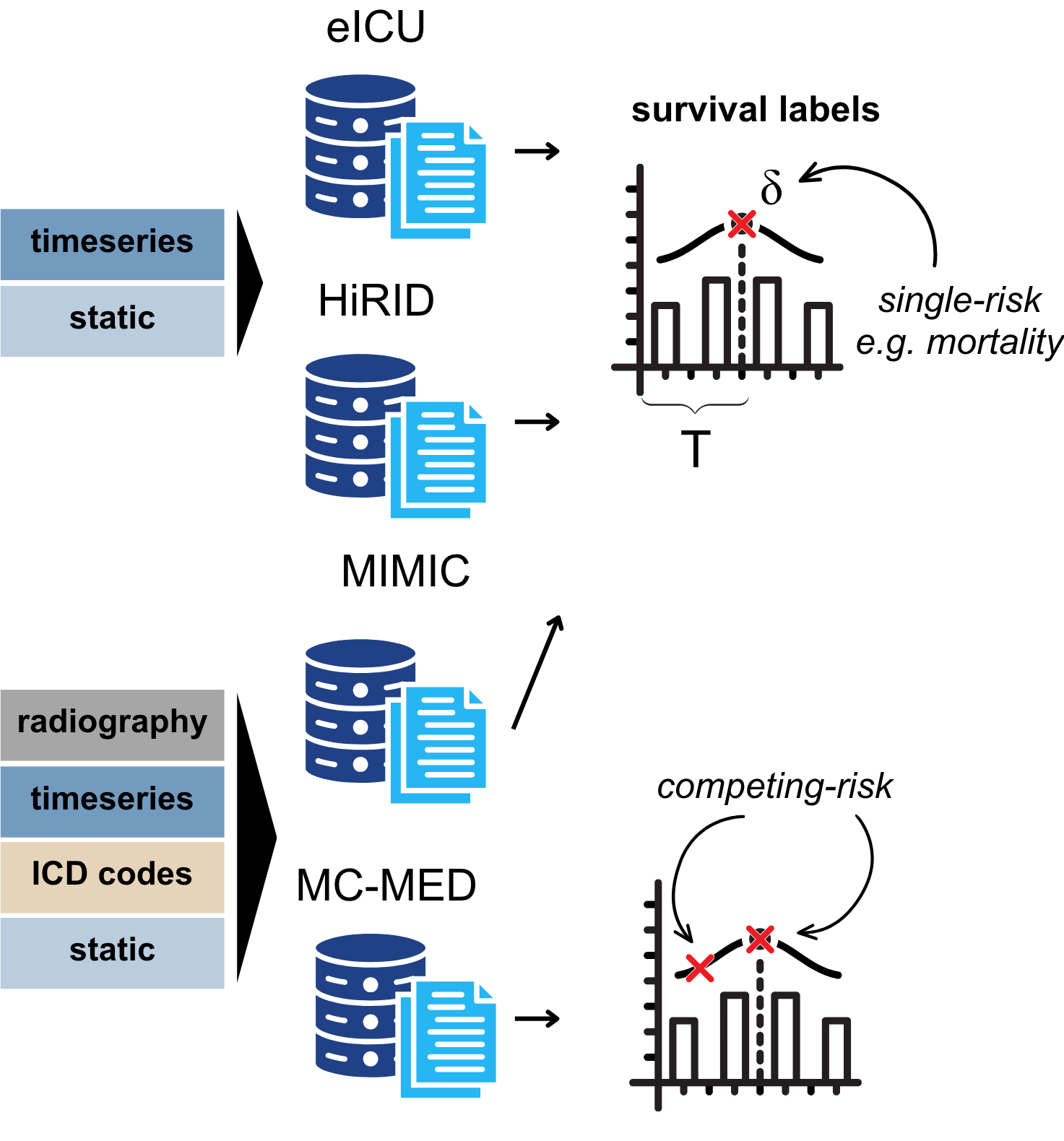}
\caption{}
\label{fig:pipeline_b_datasets}
\end{subfigure}

\vspace{0.7em}

\begin{subfigure}[t]{0.48\textwidth}
\centering
\includegraphics[width=\textwidth]{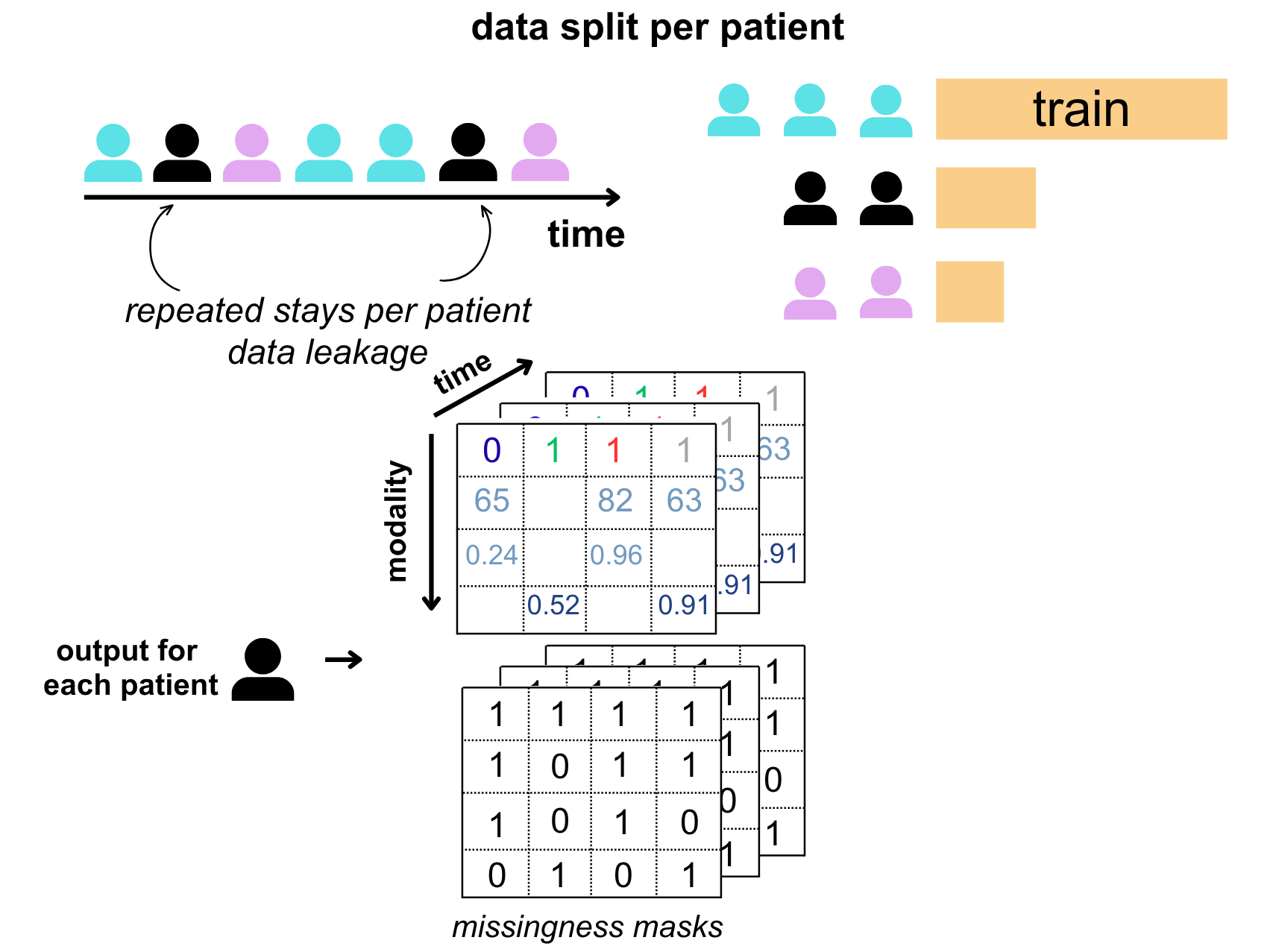}
\caption{}
\label{fig:pipeline_c_splitting}
\end{subfigure}
\hfill
\begin{subfigure}[t]{0.48\textwidth}
\centering
\includegraphics[width=\textwidth]{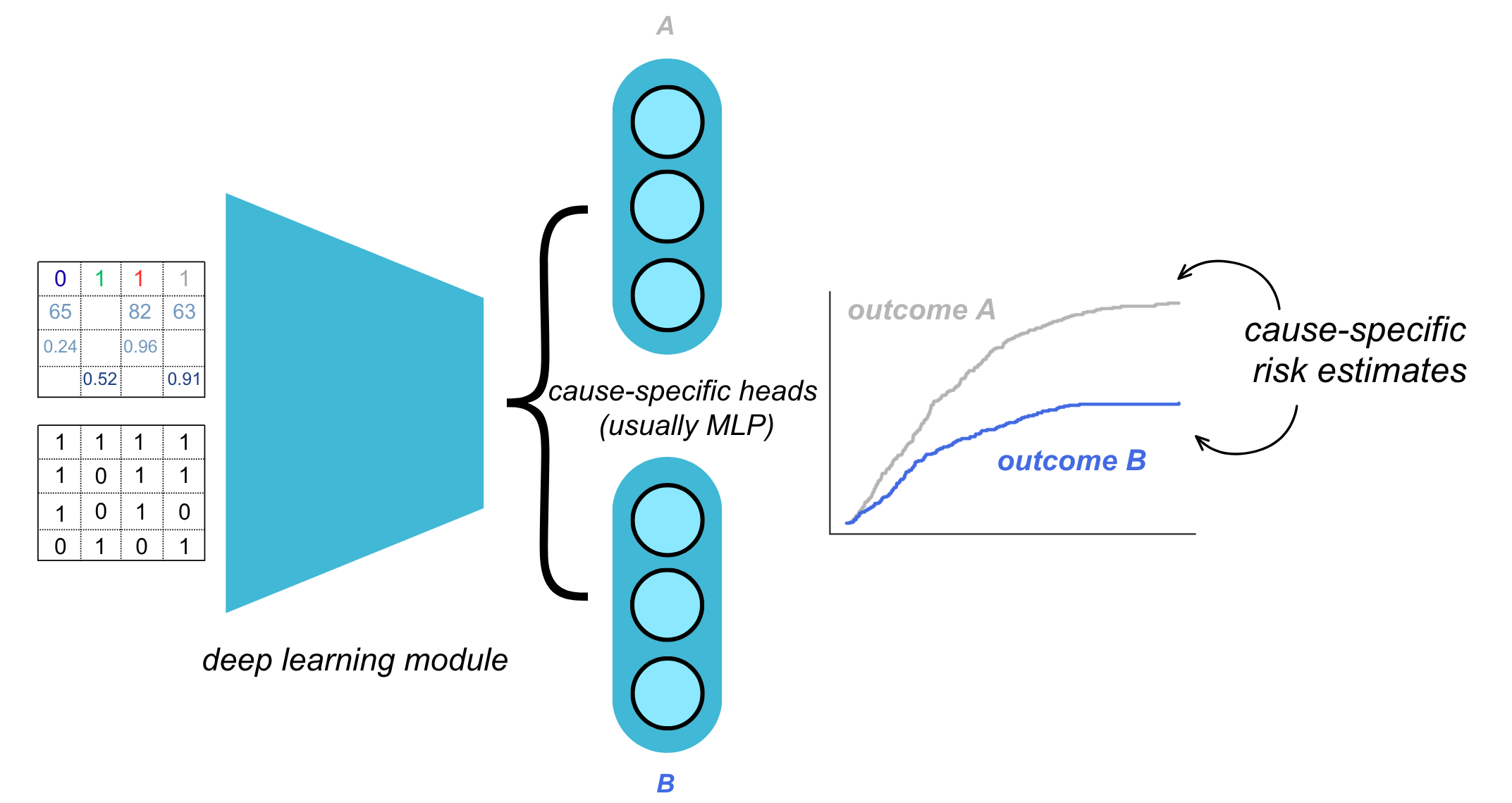}
\caption{}
\label{fig:pipeline_d_modelling}
\end{subfigure}

\caption{Overview of the SurvBench preprocessing and modelling pipeline. \textbf{(a)} The four supported input modalities: static features (age, sex, deprivation index, BMI), ICD diagnosis codes, time-series vitals and labs (e.g.\ SpO\textsubscript{2}, respiratory rate, heart rate), and radiology images with their accompanying free-text reports. Each modality follows its own preprocessing path before integration. \textbf{(b)} Source datasets and survival-label types: eICU and HiRID supply time-series and static modalities for a single-risk ICU mortality endpoint, with duration $T$ and event indicator $\delta$. MIMIC-IV adds ICD codes and radiology reports to the same single-risk task, and MC-MED supplies time-series, static, and ICD modalities for a competing-risks ED admission-pathway task. \textbf{(c)} Patient-level splitting and tensor output: repeated stays from the same patient are kept within a single fold to avoid data leakage, the train/validation/test split (70/10/20) is stratified on per-patient event status, and the per-patient output is a (modality $\times$ time) feature tensor accompanied by a binary missingness mask of identical shape. \textbf{(d)} Downstream survival modelling: the feature tensor and missingness mask feed a deep-learning module followed by cause-specific heads (typically MLPs). For competing-risks tasks, each head outputs a cause-specific cumulative incidence function over the prediction horizon. Stages (a)--(c) are implemented by the pipeline. Panel (d) is illustrative of the modelling tasks the tensors enable.}
\label{fig:pipeline_overview}
\end{figure*}

SurvBench is a configurable preprocessing pipeline that builds standardised, reproducible survival-analysis cohorts from raw EHR exports. The workflow (Figure~\ref{fig:pipeline_overview}) uses data from four large public databases (MIMIC-IV, eICU, MC-MED, HiRID), processes four modalities (static, time-series, ICD codes, radiology), and produces model-ready tensors with binary missingness masks. Splits are at the patient level. Outputs feed both single-risk and competing-risks survival models. Implementation details that are not load-bearing for the methodological argument are deferred to the Supplementary Material.

\subsection*{DATA SOURCES}

SurvBench supports four publicly available critical-care databases that span ICU and ED settings. Detailed file mappings, column references, and access requirements for each are in Supplementary §S1.1.

The Medical Information Mart for Intensive Care IV (MIMIC-IV) is a deidentified database of patients admitted to Beth Israel Deaconess Medical Centre between 2008 and 2019 \citep{johnson2023mimic}. We use v3.1 and extract patient demographics, ICU admission timing, hospital discharge outcomes, vital signs, laboratory results, ICD-9 and ICD-10 diagnosis codes, and radiology reports. Access requires CITI training and a PhysioNet data use agreement.

The eICU Collaborative Research Database covers more than 200{,}000 ICU admissions across 335 hospitals in the United States, collected between 2014 and 2015 through the Philips eICU telehealth programme \citep{pollard2018eicu}. We use v2.0 and extract demographics, vital signs (high-frequency periodic and lower-frequency aperiodic measurements), and laboratory results from hundreds of distinct assays. Where MIMIC-IV is single-centre, eICU is multi-centre and supports evaluation of generalisability across clinical settings, hospital types, and geography.

MC-MED is a deidentified multi-modal dataset from the Stanford Health Care emergency department, covering 118{,}385 adult ED visits between 2020 and 2022 \citep{kansal2025mc}. We use v1.0.1 and extract demographics, triage information (chief complaint, acuity), ED disposition, minute-resolution vital signs, laboratory results, past medical history with ICD codes, home medication lists, and radiology reports. Disposition is one of four mutually exclusive outcomes: discharge home, hospital admission, observation, or ICU admission. In-ED death is rare in the source data (under 0.1\% of visits) and is encoded as a hospital admission rather than as a separate disposition. SurvBench follows the original v1.0.1 labelling.

The HiRID-ICU is a deidentified database of approximately 33{,}000 admissions to the Department of Intensive Care Medicine at Bern University Hospital, Switzerland, between 2008 and 2016, released through PhysioNet \citep{yeche2021hirid}. HiRID provides physiological measurements at 2 to 5 minute granularity, considerably finer than the hourly resolution typical of MIMIC-IV or eICU. SurvBench targets the imputed-stage release of v1.1.1, in which the publishers have already forward-filled and grid-aligned the raw measurement stream onto a regular 5-minute axis. We extract 18 dynamic features matching the HiRID benchmark conventions: 13 vital-sign and physiological signals (heart rate, arterial systolic, diastolic, and mean blood pressure, cardiac output, peripheral oxygen saturation, Richmond Agitation-Sedation Scale, peak airway pressure, arterial and venous lactate, International Normalised Ratio, glucose, C-reactive protein) and 5 pharmacological exposure indicators (dobutamine, milrinone, levosimendan, theophylline, analgesics).

\subsection*{COHORT DEFINITION}

For each dataset, SurvBench applies configurable cohort criteria specified in the YAML configuration. The full parameter list is in Supplementary §S4.

For MIMIC-IV, the cohort is built by linking ICU stays to hospital admissions. Patients under 18 years are excluded, and a 24-hour minimum length-of-stay threshold removes short stays that often represent transfers or administrative artefacts, and we follow others in this \citep{escobar2011intra, purushotham2018benchmarking, harutyunyan2019multitask}. Stays without valid admission and discharge timestamps are dropped at the cohort merge by inner-join semantics across the ICU stays, hospital admissions, and patients tables. The patient identifier is used as the split key and the stay identifier as the data tensor index. MIMIC-IV's HIPAA-compliant deidentification caps recorded age at 91 for patients aged 89 or older, and SurvBench reads this field unchanged.

For eICU, cohort selection starts from the patient table. Age filtering removes non-numeric ages and ages under 18, including the ``greater than 89'' deidentification entries. The 24-hour minimum stay applies to unit discharge offset, and missing discharge-status values are coerced to censored ($\delta = 0$) by the boolean comparison that defines the event indicator. The patient health system identifier is the split key, and the unit-stay identifier is the data index.

MC-MED is published as an adult-only cohort upstream (118{,}385 visits in v1.0.1), and SurvBench inherits this restriction without applying its own age filter. The configuration applies no minimum visit duration by default. ED visits include short presentations such as rapid triage, transfers out, and leave-without-being-seen, and we treat these as legitimate patient trajectories rather than artefacts. Researchers wanting to exclude them can raise the minimum-stay parameter in the YAML.

For HiRID, cohort selection follows the conventions of the HiRID-ICU-Benchmark release \citep{yeche2021hirid}. Each row in the patient labels table corresponds to a single ICU stay, and the dataset is one-stay-per-patient by construction, so the patient-level identifier and the stay-level identifier coincide. SurvBench joins the labels table with the general table on patient ID and drops any patient with missing age or sex (typically $\sim$99\% kept). A 24-hour minimum length-of-stay filter is applied. The single-risk event is in-ICU mortality, derived from the discharge status field: ``dead'' maps to $\delta = 1$ and all other dispositions to $\delta = 0$.

\subsection*{SURVIVAL LABEL PROCESSING}

SurvBench supports both single-risk and competing-risks scenarios \citep{austin2016introduction}. Survival labels for each cohort consist of a duration ($T$), the time from admission to event or censoring in hours, and an event indicator ($\delta$). For single-risk scenarios, the indicator is binary, $\delta \in \{0, 1\}$, with 1 the event and 0 censoring. For competing-risks frameworks, the indicator is integer-encoded, $\delta \in \{0, 1, 2, \ldots, K\}$, with 0 censoring and $k \in \{1, \ldots, K\}$ the $k$-th competing event, supporting cause-specific hazard modelling \citep{lee2018deephit}.

Censoring ($\delta = 0$) is applied in two cases. The first is standard right-censoring, where the patient's observation period ends without a recorded terminal event (for example, discharge alive). The second is administrative censoring, where a known event occurs after a pre-defined maximum prediction horizon $H$ specified in the configuration (for example, 240 hours for MIMIC-IV). In that case the event is set to $\delta = 0$ and the duration is capped at $T = H$:
\begin{equation}
T' = \min(T, H), \quad \delta' = \begin{cases} \delta & \text{if } T \leq H \\ 0 & \text{if } T > H. \end{cases}
\end{equation}

For MIMIC-IV, the event indicator is ICU mortality. $\delta = 1$ when the patient's recorded death time falls within the ICU stay (before recorded ICU outtime) and within the configured horizon, otherwise $\delta = 0$. Duration is the time from ICU intime to that death (for events) or to outtime (for censored patients), capped at $H$. Patients dying after ICU discharge are censored at outtime \citep{kvamme2019time}. For eICU, duration is the unit discharge offset (converted from minutes to hours), and the event indicator is ICU mortality from the unit discharge status field. For HiRID, duration is derived from the relative discharge time, and the event is in-ICU mortality.

MC-MED is set up as a three-way competing-risks endpoint over the non-discharge dispositions: ICU admission ($\delta = 1$), hospital (inpatient) admission ($\delta = 2$), and observation ($\delta = 3$). Patients discharged home are administratively censored at the time of discharge ($\delta = 0$). The cause-specific cumulative incidence functions estimated under this encoding are conditional on remaining at risk for admission, rather than describing absolute disposition probabilities in the full ED cohort. The clinical target is admission-pathway prediction, not full disposition prediction.

For discrete-time survival models, continuous event times are discretised into $K$ bins using quantile-based binning using pycox's discrete-time label transform \citep{kvamme2019time}. Quantiles are computed on the training fold only, and the bin boundaries are saved and applied to validation and test sets without refitting.

\subsection*{TIME-SERIES PROCESSING}

Time-series features (vitals, lab results) need careful temporal processing. SurvBench uses a windowed aggregation strategy that converts irregularly sampled measurements into fixed-width temporal windows. The pipeline extracts time-series data for the first $T_\mathrm{max}$ hours of each stay (default: 24 hours for ICU, 6 hours for ED). The extraction window is divided into $W$ non-overlapping windows of size $w$ hours each, with $W \times w = T_\mathrm{max}$. Within a window, all measurements are aggregated using the mean. This reduces dimensionality from potentially thousands of irregularly sampled points to a fixed-length sequence, smooths measurement noise, and produces a consistent temporal structure across patients despite varying sampling frequency.

A feature is kept only if it is present (non-NaN) in at least 1\% of training-set time windows (configurable via the missingness threshold). The threshold is computed on training data only to prevent leakage, then applied to all folds. After aggregation and filtering, time-series data is indexed by patient and time-window, with columns for dynamic features and values for the per-window aggregates.

The largest source files (eICU's vital-periodic, vital-aperiodic, and lab tables, and MIMIC-IV's chart-events and lab-events tables) are too large for in-memory processing on standard workstations. The pipeline streams each file in chunks, filters to cohort patients, bins to hourly intervals, and aggregates before merging using outer join and re-aggregating to the configured window size. An on-disk cache allows for reruns at different seeds skip the multi-tens-of-gigabyte CSV pass. HiRID uses an analogous two-stage cache for its imputed-stage CSV batches. Vital signs and laboratory values arrive in heterogeneous units across the source databases and at times within a single database (temperature in $^\circ$F versus $^\circ$C, weight in lb versus kg, common labs in mg/dL versus mmol/L), and the pipeline declares a canonical unit per feature and applies hand-coded converters at load time, with the conversions written to the run log so unit drift between dataset versions surfaces as a visible signal rather than a quiet distributional shift downstream. Full streaming logic, the unit-conversion mapping, and cache layout are in Supplementary §S1.4--S1.6.

\subsection*{MULTI-MODAL INTEGRATION}

Static and time-series modalities are concatenated along the feature dimension into unified tensors. Let $\mathbf{X}^{\text{static}} \in \mathbb{R}^{N \times F_s}$ denote static features for $N$ patients with $F_s$ static features, and $\mathbf{X}^{\text{dynamic}} \in \mathbb{R}^{N \times W \times F_d}$ denote time-series features. Static features are broadcast across all time windows, $\mathbf{X}^{\text{static}}_{\text{broadcast}} \in \mathbb{R}^{N \times W \times F_s}$, and the final multi-modal tensor is:
\begin{equation}
\mathbf{X} = [\mathbf{X}^{\text{dynamic}} \; | \; \mathbf{X}^{\text{static}}_{\text{broadcast}}] \in \mathbb{R}^{N \times W \times (F_d + F_s)}.
\end{equation}

Static features are extracted once per stay and broadcast across all temporal windows. Categorical variables are one-hot encoded, and HiRID's static block additionally bins the top-$k$ APACHE II and APACHE IV groups when the extended general table is present. Full static-feature handling is in Supplementary §S1.3.

For MIMIC-IV and MC-MED, ICD diagnosis histories are vectorised using multi-hot encoding over the top-$K$ codes (default 500). For MIMIC-IV, top-$K$ ranking is performed on the configured 70/10/20 training fold. For MC-MED, ICD-like codes from past medical history are filtered to noted-date-before-arrival before crosstab construction, removing post-arrival codes that would not be available at prediction time. ICD features are stored as separate tensors given their dimensionality.

For MIMIC-IV and MC-MED, free-text radiology reports are encoded with a configurable HuggingFace model (default: Clinical-Longformer \citep{li2022clinical} with a 1024-token cutoff). Any model exposing the standard last-hidden-state interface can be substituted using the YAML configuration. Substitutability is verified by an automated test that loads Bio\_ClinicalBERT \citep{ling2023bio+} and confirms shape and finiteness of the resulting embeddings, and the configuration comments document further alternatives (RadBERT \citep{yan2022radbert}, GatorTron \citep{yang2022gatortron}). Pooling is configurable (mean or CLS). Embeddings are cached on disk and stored as separate tensors. Embedding generation runs on CUDA using HuggingFace regardless of the compute backend.

\subsection*{MISSINGNESS, IMPUTATION, AND SCALING}

Before imputation, SurvBench creates explicit binary masks $\mathbf{M} \in \{0, 1\}^{N \times W \times F}$ recording which values were observed (1) and which were missing (0). Masks accompany every tensor so that downstream models can distinguish true zeros from imputed zeros.

Residual missingness is then resolved with a leakage-safe two-stage strategy. Dynamic features are forward-filled within each patient at hourly resolution, the natural default for clinical time series, where a recently observed measurement remains the best estimate of the current value until a new measurement is taken \citep{harutyunyan2019multitask}. Positions that remain missing after forward-fill (typically those at the start of a stay, before any measurement of a given feature) are filled with the per-feature mean computed on the training fold only. Static features use training-fold means directly.

After imputation, $z$-score normalisation is applied through scikit-learn's StandardScaler, fit independently on the dynamic and static feature blocks so that broadcast static features do not dominate the dynamic block's scaling statistics:
\begin{equation}
X'_{ijf} = \frac{X_{ijf} - \mu_f}{\sigma_f},
\end{equation}
where $\mu_f$ and $\sigma_f$ are the mean and standard deviation of feature $f$ in the training fold. Both scalers are fit on the training fold only, on the post-imputation tensor. Imputed positions, therefore, contribute to the location estimate $\mu_f$ at the per-feature mean (the fallback value) and slightly deflate the dispersion estimate $\sigma_f$, both effects bounded by the per-feature observation rate. The binary mask shipped alongside every tensor lets downstream models down-weight imputed positions where desired. Validation and test sets are transformed using the fitted training-fold scalers, and post-scaling values are clipped to $[-z_\mathrm{clip}, +z_\mathrm{clip}]$ with $z_\mathrm{clip} = 10$ by default.

Splits are at the patient level. Default ratios are 70\% training, 10\% validation, 20\% test, configurable using YAML. The procedure extracts unique patient identifiers, then partitions patients into three folds using scikit-learn's stratified train-test split, stratified on per-patient event occurrence (any event versus none) and seeded for reproducibility. All stays for each patient inherit the patient's fold assignment. Stratification keeps validation and test event rates close to the training event rate, which matters for stable calibration of discrete-time hazard models on small event counts. The match is tightest on the rare-event ICU configurations (within 0.1 percentage points for MIMIC-IV, eICU, and HiRID, Table~\ref{tab:dataset_summary}). On MC-MED, where the same subject identifier can appear across multiple visits, the patient-level split maps back to a slightly different per-visit event rate (40.6\%, 40.0\%, and 40.6\% for train, validation, and test, a 0.6 percentage-point gap on validation), still well inside the $\pm 2$ percentage-point tolerance.

Compute can run on CPU or GPU. The compute layer abstracts over pandas/NumPy and cuDF/cuML, with output tensors from the two paths matching to within $1 \times 10^{-5}$ on a fixture cohort, asserted by an automated parity test in the release checklist. Full backend details, including which stages run on GPU and which currently remain on CPU, are in Supplementary §S1.7--S1.8.

\subsection*{DEFAULT CHOICES AND THEIR LIMITATIONS}

The defaults shipped with SurvBench are clinically reasonable for the configured cohorts but they are not neutral. We document the main ones and the conditions under which a researcher should override them.

\paragraph{Horizon truncation.} The ICU configurations cap durations at 240 hours and the ED configuration at 24 hours. Truncation focuses the model on an actionable window and removes long-stay outliers, but it inflates the censoring rate and reshapes the event-time distribution. The 94.3\% MIMIC-IV training censoring rate (Table~\ref{tab:dataset_summary}) is shaped by the 10-day cap and the 24-hour minimum stay on top of an underlying ICU-mortality rate near 5--8\% in published critical-care benchmarks. Relaxing either policy lowers the figure modestly but does not change the order of magnitude. Override the horizon parameter for tasks where late mortality is part of the outcome of interest, such as 30-day in-hospital or 90-day post-discharge survival.

\paragraph{Fixed temporal aggregation.} Vitals and labs are mean-aggregated into six windows of four hours (or six windows of one hour for the ED setting). Mean aggregation smooths measurement noise and equalises sequence length across patients but discards within-window variability, which can itself carry signal: heart-rate volatility, intra-window swings in mean arterial pressure, episodic desaturations. Models targeting high-frequency variability should reduce the window-size parameter, or call the streaming-cache helper directly to consume measurement-resolution rows upstream of the pipeline binning.

\paragraph{Missingness threshold.} Features observed in fewer than 1\% of training-fold windows are dropped. The threshold trades dimensionality against keeping sparse but informative measurements. 1\% is permissive, and lowering it further mostly recovers tests that are effectively never measured in the cohort of interest. For disease-specific cohorts (acute kidney injury, sepsis), the threshold may need to rise so that diagnosis-specific labs do not appear as simultaneously rare and predictive.

\paragraph{Imputation.} Forward-fill within patient plus training-fold mean is leakage-safe and preserves the temporal structure that forward-fill is designed to capture. It is strictly an improvement on the previous post-scaling zero-fill default for downstream calibration, especially on heavily missing features. Researchers wanting to use the missingness mask as a hard input rather than as a soft auxiliary signal can patch the pipeline to skip the imputation stage, or zero out the mask-flagged cells back to NaN before model input. The per-cell mask is shipped alongside every feature tensor for exactly this purpose.

\paragraph{Patient-level stratified splits.} Stratifying on per-patient event occurrence keeps validation and test event rates close to the training event rate, which matters for calibrating discrete-time hazard models on small event counts. For analyses where temporal generalisation is the primary concern, a time-based split (for example, training on 2010--2015 and testing on 2016--2019) is preferable. SurvBench currently supports only patient-level random stratified splits, and adding a time-based variant is left for future work.

\paragraph{Cross-database comparability.} The four databases support related but distinct clinical questions: ICU mortality on MIMIC-IV, eICU, and HiRID, and ED admission pathway on MC-MED. They differ in horizon, censoring structure, patient mix, and clinical context. SurvBench standardises preprocessing within each setting, and results across databases should not be read as comparable performance numbers but as separate evaluations of the same model class on different tasks.

\subsection*{DATASET CHARACTERISTICS}

The processing code is publicly available through the project GitHub repository as code, configuration files, and documentation, but not as processed data, in line with PhysioNet data use agreements. Researchers with appropriate PhysioNet credentials can generate the processed datasets by running the pipeline with the provided configurations.

Table~\ref{tab:dataset_summary} summarises the four datasets processed under their default configurations. Cohorts span an order of magnitude in size, from 15{,}343 ICU stays in HiRID (single-centre, Bern) and 51{,}838 in MIMIC-IV (single-centre, Boston) up to 127{,}977 in eICU (multi-centre, 335 hospitals) and 106{,}610 ED visits in MC-MED (single-centre, Stanford). The three ICU configurations target a single-risk 240-hour mortality endpoint, with training event rates between 4.3\% (eICU) and 7.5\% (HiRID). MC-MED targets a 24-hour competing-risks endpoint over three competing dispositions, with discharge home as administrative censoring. The training event rate (any admission) is 40.6\%. The 70/10/20 train/validation/test split is patient-level and stratified on per-patient event status. The three ICU datasets, with one stay per patient (HiRID by construction, and MIMIC-IV and eICU after the cohort filter keeps one stay per subject), show validation and test event rates within 0.1 percentage points of the training rate. MC-MED has multiple visits per subject, so the patient-level split produces a small visit-level event-rate drift (validation 40.0\% versus training 40.6\%), within the $\pm 2$ percentage-point tolerance targeted by the pipeline.

Temporal parameters reflect each setting. The ICU datasets use a 24-hour input window divided into six 4-hour bins, and MC-MED uses a 6-hour window divided into six 1-hour bins. All four use ten quantile-based output bins for discrete-time hazard models. The feature space is multi-modal across all cohorts: each includes dynamic vitals and labs plus static demographics, with totals ranging from 35 features in HiRID (18 dynamic, 17 static) up to 115 in eICU (100 dynamic, 15 static). MIMIC-IV and MC-MED additionally provide ICD diagnosis codes (top 500 by training-fold frequency) and 768-dimensional Clinical-Longformer radiology report embeddings. Dynamic-cell missingness ranges from 0.0\% in HiRID (the imputed-stage release is grid-aligned at source) and 7.2\% in MIMIC-IV up to 55.1\% in MC-MED and 62.8\% in eICU. The difference between dense and sparse datasets reflects sampling protocol rather than data quality, and is recorded explicitly through the binary missingness masks that ship alongside every feature tensor.

\begin{table}[!ht]
\centering
\scriptsize
\setlength{\tabcolsep}{4pt}
\caption{Statistical summary of the four datasets supported by SurvBench. Total stays $N$ is the sum of train, validation, and test splits. Event rates and durations are computed on the training fold. Single-risk datasets report 240\,h ICU mortality. MC-MED reports a 24\,h emergency-department competing-risks task. Missingness is the fraction of dynamic cells that were imputed.}
\label{tab:dataset_summary}
\begin{tabular}{@{}l l rrrr@{}}
\toprule
\textbf{Category} & \textbf{Statistic} & \textbf{MIMIC-IV} & \textbf{eICU} & \textbf{HiRID} & \textbf{MC-MED} \\
\midrule
\multirow{6}{*}{Source} & Version & v3.1 & v2.0 & v1.1.1 & v1.0.1 \\
& Setting & ICU, 1 site & ICU, 335 sites & ICU, 1 site & ED, 1 site \\
& Site & BIDMC & Philips eICU & Bern Univ.\ Hosp. & Stanford ED \\
& Years & 2008--19 & 2014--15 & 2008--16 & 2020--22 \\
& Outcome & ICU mortality & ICU mortality & ICU mortality & ED disposition (CR) \\
& Reference & \citep{johnson2023mimic} & \citep{pollard2018eicu}& \citep{yeche2021hirid} & \citep{kansal2025mc} \\
\midrule
\multirow{8}{*}{Cohort} & Task type & Single & Single & Single & Competing (3) \\
& Total ($N$) & 51{,}838 & 127{,}977 & 15{,}343 & 106{,}610 \\
& Train ($n$) & 36{,}286 & 89{,}573 & 10{,}740 & 74{,}572 \\
& Val ($n$) & 5{,}184 & 12{,}813 & 1{,}534 & 10{,}766 \\
& Test ($n$) & 10{,}368 & 25{,}591 & 3{,}069 & 21{,}272 \\
& Event rate train/val/test & 5.7 / 5.7 / 5.7\% & 4.3 / 4.3 / 4.3\% & 7.5 / 7.6 / 7.5\% & 40.6 / 40.0 / 40.6\% \\
& Median dur., all (train) & 56.5\,h & 55.8\,h & 56.8\,h & 5.4\,h \\
& Median dur., censored (train) & 55.9\,h & 55.2\,h & 56.9\,h & 4.4\,h \\
\midrule
\multirow{3}{*}{Filters} & $\mathrm{age} \geq$ & 18 & 18 & -- & -- \\
& $\mathrm{age} \leq$ & 120 & 89 & 120 & 120 \\
& $\mathrm{LOS} \geq$ & 24\,h & 24\,h & 24\,h & 0\,h \\
\midrule
\multirow{5}{*}{Temporal} & Input window $T_{\max}$ & 24\,h & 24\,h & 24\,h & 6\,h \\
& Horizon $H$ & 240\,h & 240\,h & 240\,h & 24\,h \\
& Windows $W$ & 6 & 6 & 6 & 6 \\
& Window size $w$ & 4\,h & 4\,h & 4\,h & 1\,h \\
& Discrete bins ($K$) & 10 & 10 & 10 & 10 \\
\midrule
\multirow{6}{*}{Features} & Modalities & TS, S, ICD, R & TS, S & TS, S & TS, S, ICD, R \\
& Total ($F$) & 64 & 115 & 35 & 95 \\
& \quad Dynamic & 24 & 100 & 18 & 72 \\
& \quad Static & 40 & 15 & 17 & 23 \\
& ICD vocab. & 500 & -- & -- & 500 \\
& Rad.\ embedding dim & 768 & -- & -- & 768 \\
\midrule
\multirow{2}{*}{Quality} & Dyn.\ missingness & 7.2\% & 62.8\% & 0.0\% & 55.1\% \\
& Z-clip & $\pm$10 & $\pm$10 & $\pm$10 & $\pm$10 \\
\bottomrule
\end{tabular}
\end{table}

All preprocessing decisions are controlled using YAML. Path parameters specify the raw data location and processed-data destination. Temporal-window configuration includes the input window length, the number of windows, and the window size. Survival parameters include the maximum prediction horizon for outcome truncation, the number of discrete time bins, and the discretisation method for bin boundary computation. Per-modality standardisation (separate StandardScaler fits for the dynamic and static blocks on the training fold) is fixed by default, with users able to disable scaling for ablation studies and to tune the post-scaling clip threshold. Modality selection occurs through Boolean flags. Splits are configured via the train/validation/test proportions (default 0.70 / 0.10 / 0.20) and a random seed. The full configurable parameter list and the corresponding fixed pipeline invariants are in Supplementary §S4.

\subsection*{VALIDATION}

To validate the correctness of the preprocessing pipeline and the quality of the generated datasets, we provide automated validation scripts and visualisation tools.

\begin{figure}[!ht]
\centering
\includegraphics[width=\textwidth]{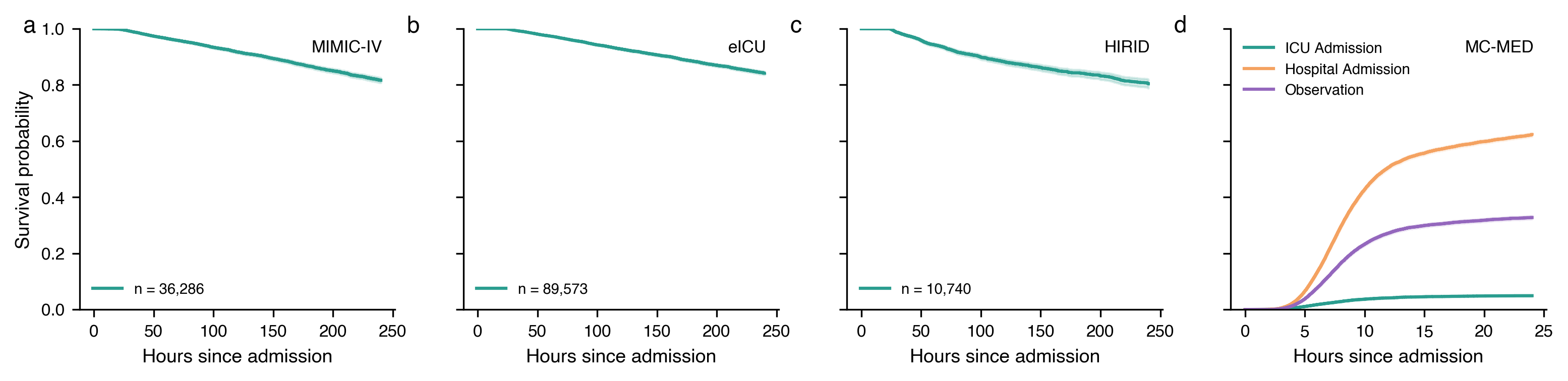}
\caption{Survival functions across the four datasets. Panels (a, b, c) show Kaplan-Meier estimates of survival probability for the single-risk ICU mortality task on MIMIC-IV, eICU, and HiRID, with 95\% confidence bands and a horizon of 240 hours. Panel (d) shows Aalen-Johansen \citep{austin2024accounting} cause-specific cumulative incidence functions for the MC-MED admission-pathway task over a 24-hour horizon, with three competing outcomes (hospital admission, observation, ICU admission). Patients discharged home are administratively censored, so the curves in panel (d) are conditional on remaining at risk for admission.}
\label{fig:survival_curves_all}
\end{figure}

Figure~\ref{fig:survival_curves_all} shows the survival functions for all four datasets across their respective horizons. The three single-risk panels (a, b, c) show monotonically decreasing Kaplan-Meier estimates \citep{dudley2016introduction}, ending at approximately 0.82 for MIMIC-IV, 0.84 for eICU, and 0.80 for HiRID. These terminal probabilities are consistent with training-fold event rates of 5.7\%, 4.3\%, and 7.5\% reported in Table~\ref{tab:dataset_summary}. The MC-MED panel (d) shows cause-specific cumulative incidence for the three competing outcomes. Hospital (inpatient) admission, $\delta = 2$, is the dominant non-discharge outcome, reaching $\sim$0.62 cumulative incidence by the 24-hour horizon. Observation, $\delta = 3$, is an intermediate disposition at $\sim$0.32. ICU admission, $\delta = 1$, is the rarest of the three at $\sim$0.05. Patients discharged home are administratively censored at the time of discharge rather than modelled as a fourth competing event, so the cumulative incidence functions in panel (d) are conditional on remaining at risk for admission and do not describe absolute disposition probabilities in the full MC-MED cohort. See SURVIVAL LABEL PROCESSING above for the methodological rationale.

\begin{figure}[!ht]
\centering
\includegraphics[width=\textwidth]{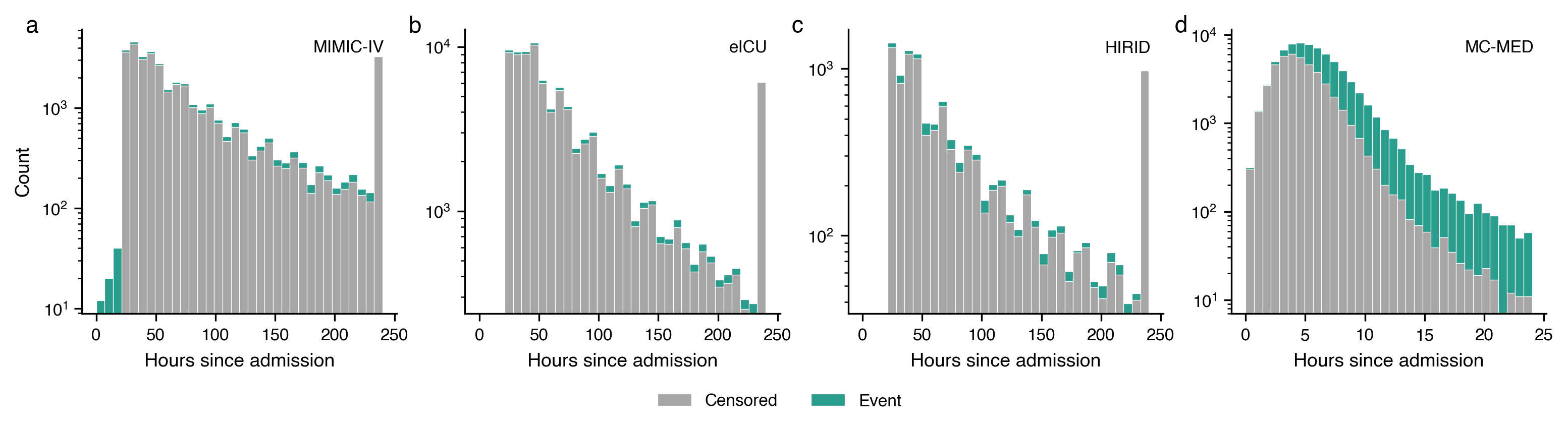}
\caption{Distribution of event and censoring times across the four datasets, log-scaled $y$-axis. Censored durations (grey) and event durations (teal) are stacked. Panels (a, b, c) span the 240-hour ICU horizon for MIMIC-IV, eICU, and HiRID, and panel (d) spans the 24-hour MC-MED horizon.}
\label{fig:duration_histograms}
\end{figure}

Figure~\ref{fig:duration_histograms} shows the distribution of event and censoring times. All three single-risk panels show terminal censoring spikes at the 240-hour horizon, accounting for 8.6\% of MIMIC-IV, 6.4\% of eICU, and 8.7\% of HiRID training-fold patients. These represent the administrative right-censoring of survivors still in the ICU at the horizon. The leftmost MIMIC-IV bins (0--20\,h) contain 57 deaths from the MIMIC-IV training fold whose recorded death time falls within the first 24\,h of the ICU stay despite an ICU outtime of $\geq 24$\,h passing the cohort filter. This dual-source discrepancy is specific to MIMIC-IV's split-table architecture. By contrast, eICU and HiRID draw both timestamps from a single source and contain zero patients with duration below 20\,h after the same length-of-stay filter. The right-skewed shape of MC-MED (d), peaking at 4--6 hours, reflects ED workflow timescales, with most dispositions in the first half of the 24-hour horizon.

\begin{figure}[!ht]
\centering
\includegraphics[width=\textwidth]{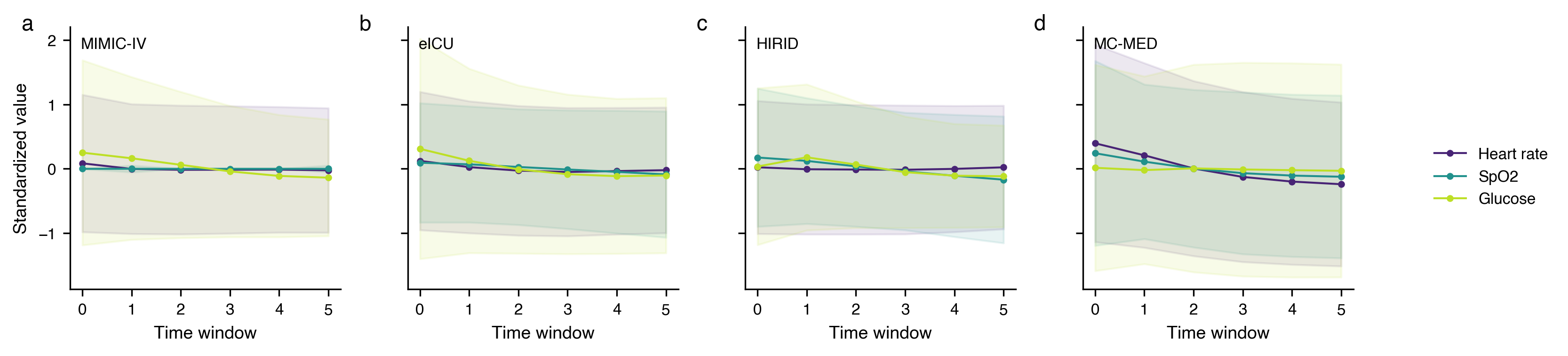}
\caption{Mean standardised trajectories of three representative dynamic features (heart rate, peripheral oxygen saturation [SpO\textsubscript{2}], glucose) across the input observation windows for each dataset. Lines show the per-window mean computed from observed values only (using the missingness masks). Shaded bands show $\pm$1 standard deviation. Trajectories are reported on the post-scaling $z$-score axis.}
\label{fig:feature_trajectories}
\end{figure}

Figure~\ref{fig:feature_trajectories} shows mean trajectories ($\pm$1 SD) for three representative dynamic features (heart rate, SpO\textsubscript{2}, glucose) across the input observation windows of each dataset. Standardised values ($z$-scores) confirm scaling worked: features are centred near zero across all four panels. The width of the standard-deviation bands tracks dataset-level missingness: narrow bands in MIMIC-IV (a, 7.2\% missing), wider in eICU (b, 62.8\%) and MC-MED (d, 55.1\%), intermediate in HiRID (c, 0.0\% post-imputation but with smoothing inherited from the imputed-stage release). Heart rate and SpO\textsubscript{2} are stable across windows in all four datasets, consistent with continuous monitoring and homeostatic regulation. Glucose drifts modestly downwards in MIMIC-IV and eICU, consistent with routine glycaemic management \citep{nice2009intensive}. The MC-MED heart-rate trajectory (d) declines over the 6-hour input window, probably from the ED admission pattern of tachycardia at presentation, resolving as treatment begins.

The pipeline enforces three correctness mechanisms at preprocessing time. Schema validation against the per-dataset schema is invoked after every CSV read, raising on missing required columns and tolerating extras. HiRID's pre-imputed time series are passed through clinical-range bounds (for example, heart rate $\in [20, 250]$\,bpm, mean arterial pressure $\in [20, 200]$\,mmHg) that log and clip outliers rather than dropping rows. A post-StandardScaler $\pm 10$ $z$-clip caps heavy-tailed values on the dynamic and static feature blocks. Other correctness properties hold by construction rather than runtime assertion. Output tensors have shape $(N \times W \times F)$ from the aggregator's concatenation. Missingness masks are binary because they are built from a NaN check. Durations are clipped to the configured horizon $H$ during label generation. Event indicators take values in $\{0, \ldots, R\}$, that is, $\{0, 1\}$ for single-risk and $\{0, 1, 2, 3\}$ for MC-MED, where 0 codes both the discharge-home reference and any administrative censoring at the 24-hour horizon, and 1, 2, 3 code the three admission-type outcomes. A companion script prints per-split summaries (cohort sizes, outcome distributions, tensor shapes, missingness rates) for post-run verification. The full schema, HiRID range bounds, and additional missing-data analytics are in Supplementary §S2.


\section*{Main Results}

To exercise every modality the pipeline produces and to give downstream users a starting point, we trained five survival models on the four SurvBench cohorts. The numbers below are reference performance and are intended as a substrate against which future methods can be compared under identical preprocessing.

\subsection*{MODELS}

We selected four published baselines spanning classical and recent deep-learning survival approaches, plus one new architecture of our own. \textbf{Cox proportional hazards} \citep{fox2002cox} is fit with lifelines on mean-pooled static features. \textbf{DeepHit} \citep{lee2018deephit} is a discrete-time competing-risks neural network that takes static features and produces a joint event-time PMF using flat-softmax. \textbf{Dynamic-DeepHit} \citep{lee2019dynamic} extends DeepHit with an LSTM and temporal-attention head over the full time-series tensor, plus an auxiliary next-step prediction loss. \textbf{DySurv} \citep{mesinovic2026dysurv} is an LSTM-variational-autoencoder with cause-specific logistic-hazard heads. We add \textbf{TransformerSurv}, a pre-LN Transformer encoder over the full tensor with cause-specific heads and a DeepHit-style loss. It is designed for multi-modal inputs and trained in two configurations, one over time-series and static features only, and one over the full multi-modal stack (time-series, static, ICD, radiology) where available. Architectural and training details for all five models are in Supplementary §S5.

\subsection*{EVALUATION}

For each (model, dataset, seed) cell, we compute four metrics on the held-out test split: Antolini's time-dependent concordance ($C^{\mathrm{td}}$) \citep{antolini2005time}, the integrated Brier score (IBS) \citep{haider2020effective}, the integrated negative binomial log-likelihood (IBLL), and mean cumulative dynamic AUC \citep{lambert2016summary}. Concordance, IBS, and IBLL are computed using pycox's evaluation utilities. Cumulative dynamic AUC uses scikit-survival with Uno-style inverse-probability-of-censoring weighting from the training fold. Time points for the AUC are anchored at a fixed clinical grid, $\{24, 72, 168, 240\}$ hours for the 240-hour ICU horizon and $\{6, 12, 18, 24\}$ hours for the 24-hour MC-MED horizon, and reported alongside their integrated mean. For the competing-risks MC-MED setting, all four metrics are computed per risk by treating non-target events as competing-event-censored and using $\hat{S}_k(t) = 1 - \widehat{\mathrm{CIF}}_k(t)$ as the per-risk survival estimate. Full metric definitions, the censoring-weight construction, and per-risk handling are in Supplementary §S6.

\subsection*{TRAINING PROTOCOL}

All deep-learning models are trained for up to 100 epochs with early stopping on validation loss and a fixed random seed shared with the splitting stage. Each cell is run for 5 seeds, and we report the mean and standard deviation across seeds. Model-specific optimiser, learning-rate, and batch-size choices are documented per model in Supplementary §S5. The reported numbers use uniform class weights.

\subsection*{RESULTS}

\begin{figure}[!ht]
\centering
\includegraphics[width=\textwidth]{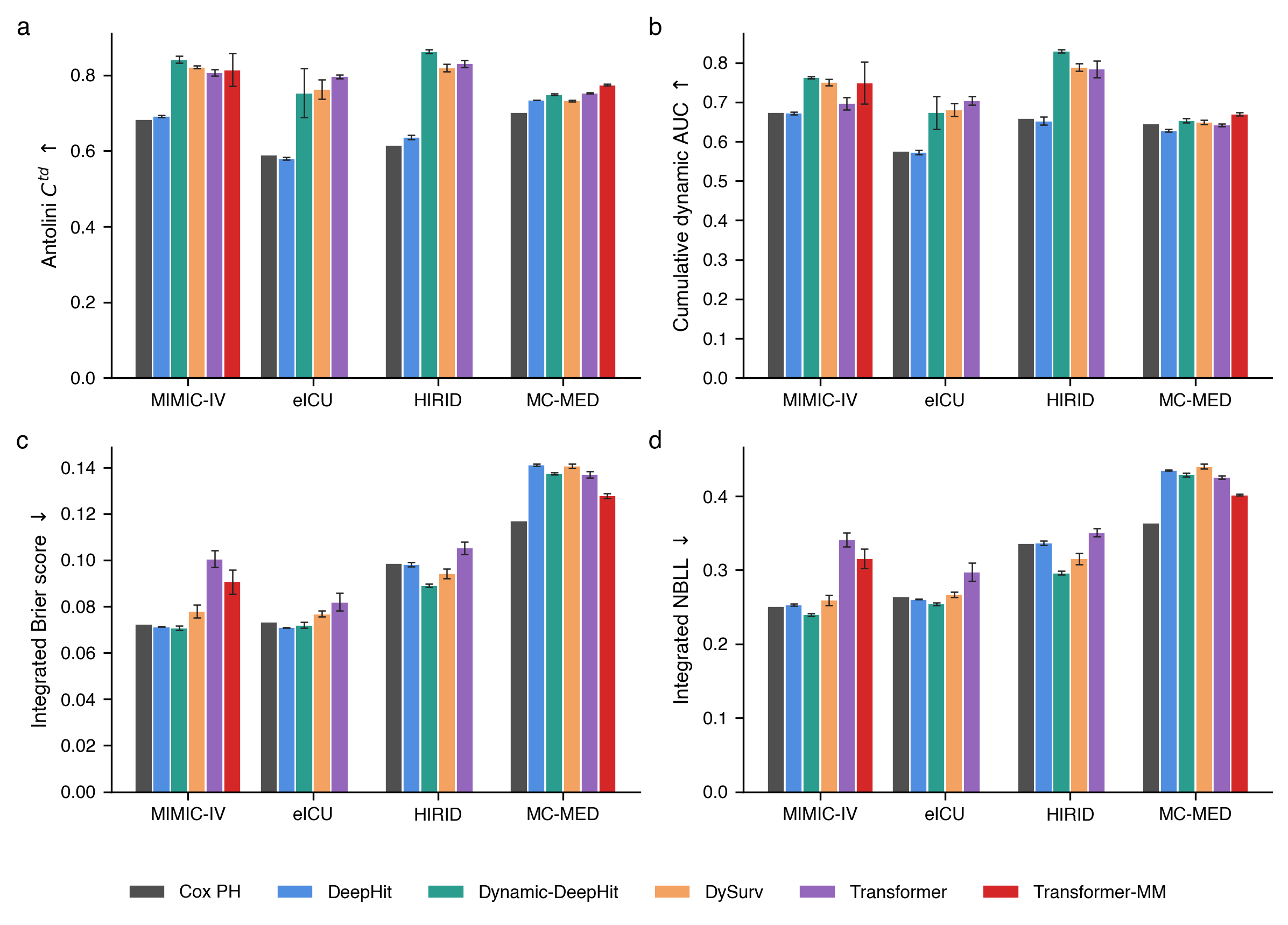}
\caption{Test-set performance of the five reference baselines across the four SurvBench cohorts, mean over 5 seeds with $\pm 1$ SD error bars. Each panel reports one metric, with one bar per model and one group of bars per dataset. \textbf{(a)} Antolini's time-dependent concordance $C^{\mathrm{td}}$ (higher is better). \textbf{(b)} Cumulative dynamic AUC integrated over the clinical grid (\{24, 72, 168, 240\}\,h for the ICU datasets, and \{6, 12, 18, 24\}\,h for MC-MED), higher is better. \textbf{(c)} Integrated Brier score (lower is better). \textbf{(d)} Integrated negative binomial log-likelihood (lower is better). TransformerSurv (multi-modal) appears only on MIMIC-IV and MC-MED, the two cohorts with ICD and radiology modalities. Cox proportional hazards is deterministic, conditional on the training fold and is shown without error bars.}
\label{fig:baseline_results}
\end{figure}

Test-set performance for all four metrics is in Figure~\ref{fig:baseline_results}. Full metric tables for all four datasets and time-anchored cumulative dynamic AUC values at each supported point of the clinical grid are in Supplementary Tables~S5--S8.

Three patterns are worth flagging, with the caveat that they hold for this specific configuration of preprocessing and these specific reference baselines and should not be read as universal claims. First, the temporal block matters. The two static-only models, Cox proportional hazards and DeepHit, sit at the bottom of every $C^{\mathrm{td}}$ column across the three single-risk ICU tasks and clear them by a wide margin on the dense-signal HiRID cohort (0.61 and 0.64 versus 0.82--0.86 for the temporal models). DeepHit is nominally a competing-risks deep network but is given only mean-pooled static features here in line with the original publication, and it lands within rounding distance of Cox on $C^{\mathrm{td}}$ and IBS in three of four datasets. The deep architecture does not in itself recover the temporal signal that windowed vitals carry. Second, multimodality helps where the modalities exist. On MC-MED, where ICD codes and radiology embeddings are available, TransformerSurv (multi-modal) is the highest-concordance model (0.774) and improves on the time-series-only TransformerSurv on both calibration metrics (IBS 0.128 versus 0.137, IBLL 0.402 versus 0.426). On MIMIC-IV, the multi-modal variant shows a similar calibration gain (IBS 0.091 versus 0.100) at the cost of variance (SD 0.044 on $C^{\mathrm{td}}$, against 0.009 for the time-series-only variant), which we attribute to small-event-count instability under the 240-hour cap and 5.7\% training event rate. Third, calibration and discrimination are separated on MC-MED. Cox proportional hazards is best on IBS (0.117) and IBLL (0.363) by a margin but worst on $C^{\mathrm{td}}$ (0.702), while TransformerSurv (multi-modal) is best on $C^{\mathrm{td}}$ but mid-pack on calibration. This is the kind of trade-off the calibration handling described in §S7 is designed to expose: a model can rank patients well while producing miscalibrated survival probabilities, and the headline number depends on which property the downstream task needs. Within each dataset, gaps between methods are smaller than the dataset-level shift in event distributions, so model choice and preprocessing both matter, and these numbers should be read as starting points for new methods rather than as a competitive leaderboard. Full provenance for every reported cell, including SurvBench commit, training-fold sample count, optimiser settings, wall time, and parameter count, is recorded in the per-run JSONs accompanying the released baseline results. Details are in Supplementary §S8.

We additionally prepared cross-dataset compatibility into the pipeline, allowing users to train on eICU and test externally on MIMIC-IV using a shared feature set (Table S9). The results of the models on this transfer task are: the Transformer achieves the highest out-of-distribution discrimination (C$^{td}$ = 0.812 $\pm$ 0.007, AUC$_{int}$ = 0.732 $\pm$ 0.009), followed by Dynamic-DeepHit (C$^{td}$ = 0.800, AUC$_{int}$ = 0.727 $\pm$ 0.003) and DySurv (C$^{td}$ = 0.799 $\pm$ 0.008, AUC$_{int}$ = 0.721 $\pm$ 0.003). The integrated  Brier score for these three temporal models lies between 0.07 and 0.08 across all five seeds. As expected, the static-only Cox and DeepHit baselines, which rely solely on the two shared demographics (age and sex), plateau near C$^{td}$ around 0.57.


\section*{Discussion}

SurvBench, for the first time, standardises the preprocessing step in deep-learning EHR survival analysis, which is the step where most cross-paper variation originates. The pipeline produces reproducible, configurable cohorts on four public databases under a single end-to-end YAML, with all leakage-prone choices (imputation, scaling, feature filtering) confined to the training fold and a binary missingness mask shipped alongside every feature tensor. The reference baselines exercise every modality the pipeline produces and offer fixed numbers that future methods can be measured against under matched preprocessing.

The four datasets target related but distinct clinical questions. ICU mortality on MIMIC-IV, eICU, and HiRID is a single-risk endpoint over a 240-hour horizon. The ED admission pathway on MC-MED is a three-way competing-risks endpoint over 24 hours. SurvBench standardises preprocessing within each task so that within-task comparisons are well defined and meaningful.

Existing foundational benchmark work on large critical care data does not address time-to-event labels with censoring, and was built on the legacy MIMIC-III, now superseded by MIMIC-IV. It also predates the recent competing-risks and multi-modal survival models that SurvBench is designed to support. To our knowledge, no public preprocessing pipeline of comparable scope exists for MIMIC-IV, eICU, MC-MED, or HiRID.

All four cohorts, however, are derived from single-region or US/European populations, and bias inherited from each source carries through the pipeline. While SurvBench standardizes preprocessing across diverse modalities and care settings, it inherits the demographic limitations of its source databases. All four cohorts (MIMIC-IV, eICU, MC-MED, HiRID) originate from high-income countries (the US and Switzerland), heavily weighting the benchmark toward well-resourced tertiary care centers and Western baseline health profiles. Furthermore, the datasets reflect localized healthcare system dynamics, such as US insurance disparities or specific institutional triage guidelines, that could influence survival trajectories and competing clinical dispositions. Consequently, while SurvBench provides a rigorous technical foundation for model comparison, performance on these benchmarks should not be conflated with readiness for global clinical deployment, particularly in under-represented, diverse, or resource-limited settings. Patient-level random splits are appropriate for calibrating discrete-time hazard models, but a time-based split would be preferable for analyses where temporal generalisation is the primary concern. SurvBench does not currently implement time-based splitting.

Future work includes time-based split support, broader GPU coverage of the loader-stage aggregation currently still on CPU, and incorporation of additional public databases as they become available. Long-term usability depends on continued compatibility with EHR database updates. Each loader runs schema validation against pinned versions at load time (MIMIC-IV v3.1, eICU v2.0, MC-MED v1.0.1, HiRID v1.1.1) and tolerates additive upstream column changes, so a new database release does not silently break a configuration. Supported version pins per release are tracked in the repository, and the GPU parity test is part of the release checklist. Ultimately, standardized pipelines like SurvBench do more than ensure algorithmic fairness and reproducibility and they help foster a healthier, more integrated research culture. By eliminating the need for engineers to endlessly reverse-engineer raw data exports, SurvBench allows interdisciplinary teams to redirect their focus toward the patient. It provides the shared infrastructure necessary for computational engineers and front-line healthcare professionals to collaborate, ensuring that the next generation of survival models are not just technically sound, but co-designed to be actionable, safe, and deeply grounded in real-world bedside data.

\backmatter

\bmhead{Code availability}

The SurvBench codebase is publicly available under the MIT licence at \url{https://github.com/munibmesinovic/SurvBench}.

\bmhead{Data availability}

The four source datasets are publicly available from PhysioNet. MIMIC-IV: \url{https://physionet.org/content/mimiciv/}. eICU Collaborative Research Database: \url{https://physionet.org/content/eicu-crd/}. MC-MED: \url{https://physionet.org/content/mc-med/}. HiRID: \url{https://physionet.org/content/HiRID/}. Access requires a PhysioNet data use agreement.

\bibliography{sn-bibliography}

\bmhead{Acknowledgements}

The authors thank Max Buhlan for assistance with Figure~\ref{fig:pipeline_overview}.

\bmhead{Funding}

MM is supported by the Rhodes Trust and the EPSRC CDT Health Data Science. TZ is supported by the Royal Academy of Engineering under the Research Fellowship scheme.

\bmhead{Author contributions}

\bmhead{Competing interests}

We declare no competing interests.

\end{document}


\title[Supplementary]{Supplementary Material: SurvBench: A Standardised Preprocessing Pipeline for Multi-Modal Electronic Health Record Survival Analysis}

\author*[1]{\fnm{Munib} \sur{Mesinovic}}\email{munib.mesinovic@eng.ox.ac.uk}

\author[1]{\fnm{Tingting} \sur{Zhu}}

\affil*[1]{\orgdiv{Department of Engineering Science}, \orgname{University of Oxford}, \orgaddress{\city{Oxford}, \country{UK}}}

\maketitle

\section{Pipeline implementation details}

This section documents the operational details of the workflow summarised in the SurvBench section of the main text. Subsection §S1.1 lists the source files and identifier columns each loader expects. Subsections §S1.2--S1.8 describe the per-stage implementation choices in the order they execute.

\subsection{Data source file mappings}
\label{sec:supp_filemap}

This subsection lists the input files, identifier columns, and access constraints for each dataset. All files are read with strict schema validation against the pinned dataset version (§S2).

\paragraph{MIMIC-IV v3.1.}

MIMIC-IV separates hospital-level data (\texttt{hosp/}) from ICU-level data (\texttt{icu/}). The pipeline uses \texttt{icu/icustays.csv} as the temporal anchor, \texttt{hosp/patients.csv} for demographics, and \texttt{hosp/admissions.csv} for hospital-level discharge outcomes. Time-series data come from \texttt{icu/chartevents.csv} (vitals) and \texttt{hosp/labevents.csv} (laboratory results). Diagnosis codes (ICD-9 and ICD-10) are read from \texttt{hosp/diagnoses\_icd.csv}, and free-text radiology reports are read from \texttt{note/radiology.csv}, distributed under MIMIC-IV-Note as a companion release to MIMIC-IV core, when the radiology modality is enabled. Identifier columns: \texttt{subject\_id} is the patient-level split key, and \texttt{stay\_id} is the per-stay tensor index. The \texttt{anchor\_age} field is read unchanged. Access requires CITI training and a PhysioNet data use agreement.

\paragraph{eICU Collaborative Research Database v2.0.}

Pre-integrated patient records sit in \texttt{patient.csv}. Time-series data are split across three files: \texttt{vitalPeriodic.csv} (high-frequency vitals at roughly 5-minute intervals), \texttt{vitalAperiodic.csv} (lower-frequency irregular measurements), and \texttt{lab.csv} (results from hundreds of distinct assays). The pipeline excludes \texttt{nurseCharting.csv}, which is over 100\,GB uncompressed and predominantly unstructured. Identifier columns: \texttt{patienthealthsystemstayid} is the patient-level split key, and \texttt{patientunitstayid} is the per-stay tensor index. The eICU loader renames these to \texttt{subject\_id} at cohort load time so downstream code paths are dataset-agnostic.

\paragraph{MC-MED v1.0.1.}

The master file \texttt{visits.csv} contains visit identifiers, demographics, triage information (chief complaint, acuity), and ED disposition. Time-series data covers minute-resolution vitals (\texttt{numerics.csv}) and laboratory results (\texttt{labs.csv}). Historical context is provided by \texttt{pmh.csv} (past medical history with ICD codes). Free-text radiology reports are in \texttt{rads.csv}. MC-MED also distributes a \texttt{meds.csv} home medication list, but a medication modality is not implemented in the current loader. The \texttt{pmh.csv} loader applies a causal filter \texttt{Noted\_date < Arrival\_time} to remove ICD codes recorded after the visit's arrival timestamp, which would otherwise leak post-arrival information into a prediction-time feature. ED disposition in \texttt{visits.csv} encodes one of four mutually exclusive outcomes (discharge home, hospital admission, observation, ICU admission), and the loader maps this to the three-way competing-risks indicator described in the main text.

\paragraph{HIRID v1.1.1.}

The pipeline consumes \texttt{general\_table.csv} (demographics and discharge status), \texttt{benchmark\_output/hirid\_patient\_labels.csv} (per-patient survival labels), the optional \texttt{benchmark\_output/general\_table\_extended.parquet} (height, APACHE II Group, APACHE IV Group), and the directory \texttt{imputed\_stage/csv/} containing approximately 252 batched CSV files totalling around 5.5\,GB. Identifier column: \texttt{patientid} serves both as the patient-level split key and as the per-stay tensor index, since HIRID is one-stay-per-patient by construction.

\subsection{Stage 1: Cohort loading}

For eICU, we use a memory-efficient pre-aggregation strategy for the large \texttt{lab.csv}, \texttt{vitalPeriodic.csv}, and \texttt{vitalAperiodic.csv} files. Rather than loading entire files into memory and then filtering, the pipeline:

\begin{enumerate}
    \item Identifies cohort patient identifiers from \texttt{patient.csv}.
    \item Sequentially loads each large file in chunks.
    \item Filters to cohort patients and relevant time windows.
    \item Aggregates to hourly bins using groupby operations.
    \item Releases memory before loading the next file.
    \item Merges the pre-aggregated hourly data.
\end{enumerate}

This avoids the out-of-memory errors (exit code 137) that occurred in naive implementations attempting to load and merge gigabyte-scale files at once.

\subsection{Stage 4: Static feature processing}

Static features capture demographics and admission characteristics that remain constant during a hospital encounter. They are extracted once per stay and broadcast across all temporal windows so they can feed recurrent and temporal-convolutional architectures. Features typically include age (continuous) and categorical variables such as gender, ethnicity, and admission type or unit, all one-hot encoded into binary indicator columns. An ethnicity variable with four categories produces four binary features, letting the model learn ethnicity-specific patterns without imposing an ordinal relationship.

Feature filtering handles rare categorical values that appear infrequently across the cohort and may cause overfitting or numerical instability. Categories present in fewer than a configurable percentage of training patients (default 1\%) can be merged into an ``Other'' category, avoiding the high-dimensional sparse feature spaces that hurt training. The threshold is computed on training data only, with the resulting category definitions applied to validation and test.

\subsection{Stage 5: Memory-efficient streaming for eICU and MIMIC-IV}

eICU's large time-series files (\texttt{vitalPeriodic}, \texttt{vitalAperiodic}, \texttt{lab}) need special handling beyond the chunked loading in Stage 1. The pipeline:

\begin{enumerate}
    \item Loads \texttt{vitalPeriodic.csv}, filters to cohort and time range, bins measurements to hourly intervals using \texttt{observationoffset // 60}, and aggregates each hour with \texttt{groupby().mean()}.
    \item Repeats the process for \texttt{vitalAperiodic.csv}.
    \item Pivots \texttt{lab.csv} so lab names become columns (\texttt{pivot\_table}), then bins and aggregates by hour.
    \item Merges the three hourly dataframes using outer joins.
    \item Re-aggregates from hourly to 4-hour windows (or another configured window size).
\end{enumerate}

This two-stage aggregation (raw $\to$ hourly $\to$ windows) keeps memory usage bounded while preserving temporal patterns. The same logic, factored into \texttt{preprocessing/chartevents\_cache.py}, applies to MIMIC-IV \texttt{chartevents} and \texttt{labevents} streaming, with an on-disk cache so reruns at different seeds skip the multi-tens-of-gigabyte CSV pass.

\subsection{Stage 5: Unit standardisation}

Vital signs and laboratory values arrive in heterogeneous units across the source databases and at times within a single database. Temperature is recorded in $^\circ$F or $^\circ$C, weight in lb or kg, and common labs in mg/dL or mmol/L. To remove this source of distributional drift before any downstream modelling, \texttt{preprocessing/units.py} declares a canonical unit per feature and applies hand-coded converters at load time, with the conversions written to \texttt{unit\_conversions\_<prefix>.json} in the output directory. The audit log records the count of rows converted per feature kind, so unit drift between dataset versions surfaces as a visible signal rather than a silent distributional shift downstream. For MIMIC-IV chartevents, the loader normalises the \texttt{valueuom} field (lowercasing, whitespace stripping, and mapping the recorded ``F'' and ``deg F'' variants to a single canonical token) before consulting the converter table, which prevents the same physical unit from being treated as several distinct units due to formatting variation.

\subsection{Stage 5: HIRID two-stage caching}

HIRID's imputed-stage release is distributed as roughly 252 batched CSV files totalling around 5.5\,GB, so naive in-memory processing is impractical on standard workstations. SurvBench therefore uses a two-stage strategy specific to this dataset.

\paragraph{Stage 1.} The pipeline streams the imputed-stage CSVs once, keeps only the cohort patients and the 18 target features, bins each measurement to its hour index $\lfloor \texttt{reldatetime} / 3600 \rfloor$, applies forward-fill within each patient (never across patients), and accumulates a tensor of shape $(N_\mathrm{patients}, T_\mathrm{hours}, 18)$ together with companion arrays for length of stay, event status, static features, and patient identifiers. These artefacts are persisted as NumPy arrays under \texttt{<base\_dir>/.cache/hirid\_*}, with a JSON sidecar fingerprinting the static-feature schema so cache invalidation is automatic when the static block changes. The cache filename incorporates the configured input window $T$, so altering \texttt{preprocessing.max\_hours} produces a fresh cache rather than silently mismatched shapes.

\paragraph{Stage 2.} The pipeline reuses the cached arrays on every subsequent run, including reruns at different random seeds or with different scaling parameters, and applies the same downstream cohort filter, train/val/test split, train-fold imputation fallback, per-modality scaling, and \texttt{pycox} discrete-time bin construction as the other three datasets. First-run cost is around 30 minutes, and subsequent runs finish in under a minute.

The static feature block is enriched when \texttt{general\_table\_extended.parquet} is available. In addition to age and sex, the loader emits height (mean-imputed across the cohort) and one-hot encodings of the top-$k$ APACHE II Group and APACHE IV Group codes plus presence indicators, providing severity context that the imputed-stage time-series alone does not convey.

\subsection{Compute backend}

Compute can run on CPU or GPU. The compute layer, factored into \texttt{preprocessing/backend.py}, abstracts over pandas/NumPy and cuDF/cuML behind a single \texttt{compute.backend} setting. With \texttt{compute.backend: gpu}, the pipeline routes the chartevents and labevents streaming pass, the StandardScaler fit and transform, the train-fold imputation fallback, and tensor I/O through cuDF and cuML. The backend automatically falls back to pandas and scikit-learn when the GPU libraries are not importable, so a single configuration runs on both CPU-only and GPU-equipped machines without edits. The loader-stage pandas aggregation, including \texttt{groupby}, \texttt{pivot}, and forward-fill, currently remains on CPU and is queued for the next release. Embedding generation always runs on CUDA via HuggingFace regardless of the compute backend, because the radiology pipeline depends on a transformer model loaded under \texttt{torch.cuda}. The GPU dependencies (cuDF, cuML, RAPIDS) are pinned in a separate \texttt{requirements-gpu.txt} so the default CPU install path remains lightweight.

\subsection{GPU parity test}

The GPU compute path produces output tensors that match the CPU path to within $1 \times 10^{-5}$, asserted by \texttt{tests/test\_backend\_parity.py} via \texttt{np.testing.assert\_allclose} with \texttt{atol=1e-5}. The residual numerical gap reflects different accumulation orders between scikit-learn and cuML rather than a correctness difference between the two paths. The test runs the pipeline on a small fixture cohort under both backends and compares every output array. It skips automatically when cuDF is not installed (i.e., on CPU-only machines and continuous integration runners without GPU). It is part of the release checklist in \texttt{SUPPORTED\_VERSIONS.md}, and a release tag may not ship until the parity test passes on a machine with RAPIDS installed.

\section{Data quality checks}

The pipeline distinguishes three runtime enforcement mechanisms from properties that hold by construction.

\paragraph{Schema validation (enforced).} Each loader calls \texttt{validate\_schema()} after every \texttt{pd.read\_csv}, comparing column names against the schema declared in \texttt{data/schema.py} for the pinned dataset version. The validator checks required columns only and tolerates extras, so additive upstream changes do not break existing configurations. A missing required column raises \texttt{SchemaValidationError} listing the offending file and columns. Dtype hints are stored in the schema dataclass but are advisory only. Pandas infers dtypes loosely on read, and the loaders coerce types explicitly where needed downstream rather than asserting them at load time.

\paragraph{HIRID range bounds (enforced).} HIRID's pre-imputed time series passes through \texttt{clip\_to\_ranges()} after the Stage-1 forward-fill, which clips values to clinical bounds and writes an entry to the run log for each clipped row. Bounds are listed in Table~\ref{tab:hirid_ranges}. Values are clipped, never dropped.

\paragraph{Post-scaling $z$-clip (enforced).} After per-modality StandardScaler fit-transform on the training fold, values are clipped to $[-z_\mathrm{clip}, +z_\mathrm{clip}]$ on the dynamic and static blocks. Default $z_\mathrm{clip} = 10$, configurable via \texttt{preprocessing.scaling.z\_clip}.

\paragraph{Properties that hold by construction.} The following are true at output time but are not actively asserted:

\begin{itemize}
    \item \emph{Tensor shape} $(N \times W \times F)$: produced by the aggregator's concatenation of dynamic and static blocks.
    \item \emph{Binary missingness masks}: built as $\sim\!\texttt{np.isnan}(\mathbf{X})$, so 0/1 by construction.
    \item \emph{Non-negative durations capped at $H$}: clipped during label generation in each loader.
    \item \emph{Event-indicator range} $\{0, \ldots, R\}$: produced directly by the label-generation logic.
    \item \emph{Approximately standard-normal scaled training features}: guaranteed by \texttt{StandardScaler.fit\_transform} on the training fold (mean $\to$ 0, standard deviation $\to$ 1) modulo the $z$-clip cap. Validation and test sets may show some distributional shift reflecting temporal or population differences.
\end{itemize}

\paragraph{Post-run verification (advisory).} A companion script, \texttt{scripts/validate\_data.py}, prints per-split summaries such as cohort sizes, outcome distributions, tensor shapes, feature counts, and missingness percentages for manual review. The script does not assert. It is intended for spot-checking after a run rather than gating the run.

HIRID loading additionally applies clip-and-log range validation on continuous variables. The default clinical-range bounds, tunable in \texttt{data/schema.py}, are listed in Table~\ref{tab:hirid_ranges}. Values outside the bounds are clipped to the boundary, and an entry is written to the log. Values are not dropped. The bounds are deliberately conservative for an ICU population so that legitimate extremes (e.g., a heart rate of 240 in tachyarrhythmia) are kept rather than discarded.

\begin{table}[h]
\centering
\small
\caption{Clinical-range bounds applied as clip-and-log validation in the HIRID loader. Out-of-range values are clipped to the boundary and logged.}
\label{tab:hirid_ranges}
\begin{tabular}{lll}
\toprule
\textbf{Feature} & \textbf{Lower bound} & \textbf{Upper bound} \\
\midrule
Heart rate (bpm) & 20 & 250 \\
SpO\textsubscript{2} (\%) & 50 & 100 \\
Mean arterial pressure (mmHg) & 20 & 230 \\
Systolic blood pressure (mmHg) & 30 & 250 \\
Diastolic blood pressure (mmHg) & 10 & 200 \\
Peak airway pressure (cmH\textsubscript{2}O) & 0 & 80 \\
Glucose (mmol/L) & 1 & 50 \\
Lactate, arterial and venous (mmol/L) & 0 & 30 \\
INR & 0.5 & 20 \\
C-reactive protein (mg/L) & 0 & 600 \\
\bottomrule
\end{tabular}
\end{table}

\paragraph{Missing data analysis.}

The pipeline produces per-feature missingness statistics, including the percentage of windows containing observed values before imputation, the identification of features dropped due to excessive missingness relative to the configured threshold, and a comparison of missingness rates across folds.

\section{Usage notes}

\subsection{Installation}

\begin{enumerate}
    \item Obtain credentialed access to the desired PhysioNet datasets.
    \item Clone the repository: \texttt{git clone https://github.com/munibmesinovic/SurvBench.git}.
    \item Install dependencies: \texttt{pip install -r requirements.txt} for the CPU path, plus \texttt{pip install -r requirements-gpu.txt} for the optional GPU path.
    \item Download raw CSV files from PhysioNet to a local directory.
    \item Edit configuration files to specify data paths.
    \item Run preprocessing: \texttt{python scripts/run\_preprocessing.py --config configs/<dataset>\_config.yaml}.
\end{enumerate}

\subsection{Loading processed data}

\begin{verbatim}
import numpy as np
import pickle
from pathlib import Path

# Load feature tensors
data_dir = Path("path/to/output")
x_train = np.load(data_dir / "x_train_eicu.npy")
m_train = np.load(data_dir / "x_train_eicu_mask.npy")

# Load survival labels
with open(data_dir / "y_train_surv_eicu.p", 'rb') as f:
    durations_train, events_train = pickle.load(f)

# Load feature metadata
with open(data_dir / "feature_names_eicu.pkl", 'rb') as f:
    feature_info = pickle.load(f)

print(f"Training data shape: {x_train.shape}")
print(f"Dynamic features: {feature_info['num_dynamic']}")
print(f"Static features: {feature_info['num_static']}")
\end{verbatim}

\subsection{Configuration customisation}

Common modifications include adjusting the temporal window (\texttt{max\_hours}, \texttt{num\_windows}, \texttt{window\_size\_hours}) for different resolutions, modifying the survival horizon (\texttt{max\_horizon\_hours}) for different time scales, adjusting \texttt{missingness\_threshold} for feature-filtering aggressiveness, and toggling modality flags (\texttt{icd}, \texttt{radiology}) for ablation studies. The full list of exposed parameters is in §S4.1.

\subsection{Extending to new datasets}

\begin{enumerate}
    \item Create a new data loader class inheriting from \texttt{BaseDataLoader}.
    \item Implement abstract methods: \texttt{load\_labels()}, \texttt{load\_static\_features()}, \texttt{load\_timeseries()}, \texttt{apply\_cohort\_criteria()}.
    \item Add a schema entry in \texttt{data/schema.py} listing required columns and dtypes.
    \item Create a YAML configuration file specifying file paths and cohort criteria.
    \item Register the new loader in \texttt{pipeline.py}.
\end{enumerate}

The HIRID loader is the most recent example of this pattern.

\subsection{Common pitfalls and troubleshooting}

\paragraph{Memory limitations.} The most common technical challenge for large databases such as eICU. Mitigations include compressed CSV files (\texttt{.csv.gz}), reducing \texttt{max\_hours}, increasing system swap, processing on high-memory compute nodes (128--256\,GB RAM), and enabling the GPU backend.

\paragraph{Patient-level split failures.} These typically arise from missing or incorrectly named subject identifier columns. The pipeline requires a \texttt{subject\_id} column. Loaders must rename dataset-specific identifier columns (e.g., \texttt{patienthealthsystemstayid} in eICU) at cohort load time.

\paragraph{Feature-filtering aggressiveness.} Setting \texttt{missingness\_threshold} too high (e.g., 0.10) may drop sparsely-but-informatively measured features such as troponin, which is typically only ordered when acute coronary syndrome is suspected but whose presence and magnitude strongly predict mortality. Start at the default 0.01 and only raise it if dimensionality becomes prohibitive (e.g., over 500 features), or if pilot experiments suggest sparse features add more noise than signal.

\paragraph{Temporal alignment.} The pipeline enforces \texttt{num\_windows} $\times$ \texttt{window\_size\_hours} $=$ \texttt{max\_hours} exactly. A 48-hour observation window can be 8 windows of 6, 12 of 4, or 24 of 2, but not 7 or 13.

\subsection{Performance considerations}

Preprocessing time varies by dataset and hardware:

\begin{itemize}
    \item MIMIC-IV: 1--2 hours on a workstation with 64\,GB RAM.
    \item eICU: 2--4 hours on the same workstation, dominated by the chartevents and labevents streaming passes.
    \item MC-MED: 1--2 hours on the workstation.
    \item HIRID: $\sim$30 minutes for the first run (Stage 1 cache build), then under a minute for subsequent runs (cache reuse).
\end{itemize}

The GPU backend (§S1.7) accelerates the streaming, scaling, and imputation-fallback stages. We have not yet measured end-to-end speedup against an A100 baseline, so we do not quote one. Wall time on cohorts where the chartevents pass dominates (MIMIC-IV, eICU) is expected to drop substantially, but quantification is left to the next release once GPU benchmarks are run.

\section{Configuration and output reference}

\subsection{Configurable parameters}

Table~\ref{tab:configurable_params} lists the parameters exposed through the YAML configuration. Defaults shown reflect the shipped configurations.

\begin{table}[ht]
\centering
\small
\setlength{\tabcolsep}{4pt}
\caption{Configurable parameters in the YAML pipeline configuration. Defaults shown reflect the shipped configurations.}
\label{tab:configurable_params}
\begin{tabular}{p{0.34\textwidth} p{0.28\textwidth} p{0.30\textwidth}}
\toprule
\textbf{Parameter} & \textbf{Default} & \textbf{Notes} \\
\midrule
\texttt{cohort.min\_age} & 18 (MIMIC, eICU), not set (HIRID, MC-MED) & Lower age cutoff. HIRID applies a null-age/sex drop instead, and MC-MED is adult-only upstream. \\
\texttt{cohort.max\_age} & 120 (MIMIC, HIRID, MC-MED), 89 (eICU) & Upper age cutoff. The eICU cap of 89 is a deidentification artefact. \\
\texttt{cohort.min\_stay\_hours} & 24 (MIMIC, eICU, HIRID), 0 (MC-MED) & Minimum stay or visit duration. \\
\texttt{preprocessing.max\_hours} & 24 (ICU), 6 (ED) & Input observation window. \\
\texttt{preprocessing.num\_windows} & 6 & Must satisfy $W \times w = T_{\max}$. \\
\texttt{preprocessing.window\_size\_hours} & 4 (ICU), 1 (ED) & Aggregation bin width. \\
\texttt{preprocessing.max\_horizon\_hours} & 240 (ICU), 24 (ED) & Events after this time are right-censored. \\
\texttt{preprocessing.n\_time\_bins} & 10 & For discrete-time models. \\
\texttt{preprocessing.discretisation\_method} & \texttt{quantiles} & \texttt{quantiles} or \texttt{uniform}. \\
\texttt{preprocessing.missingness\_threshold} & 0.01 & Minimum observed fraction in training fold. \\
\texttt{preprocessing.scaling.method} & \texttt{standard} & \texttt{standard} or \texttt{none}. Disabling is intended for ablation studies. \\
\texttt{preprocessing.scaling.z\_clip} & 10.0 & Post-scaling clip. \texttt{null} disables. \\
\texttt{compute.backend} & \texttt{auto} & \texttt{auto} \textbar{} \texttt{cpu} \textbar{} \texttt{gpu}. \texttt{auto} uses GPU when RAPIDS is importable, otherwise falls back to pandas/sklearn. Routes scaling, imputation fallback, and tensor I/O. Loader-stage aggregation stays on CPU. \\
\texttt{dataset.cache\_dir} & \texttt{null} (resolves to \texttt{<base\_dir>/.cache/}) & On-disk cache for streamed CSVs and embeddings. \\
\texttt{radiology\_processing.model\_name} & Clinical-Longformer & Any HuggingFace AutoModel exposing \texttt{last\_hidden\_state}. \\
\texttt{radiology\_processing.max\_length} & 1024 & Token cutoff. \\
\texttt{radiology\_processing.batch\_size} & 32 & Per-GPU batch. \\
\texttt{radiology\_processing.pooling} & mean & \texttt{mean} or \texttt{cls}. \\
\texttt{modalities.\{timeseries,static,icd,radiology\}} & dataset-specific & Boolean toggles for ablation studies. \\
\texttt{splits.\{train,val,test\}} & 0.70 / 0.10 / 0.20 & Must sum to 1. \\
\texttt{splits.seed} & 42 & For reproducibility. \\
\bottomrule
\end{tabular}
\end{table}

\subsection{Pipeline-level invariants}

A second class of pipeline behaviour, listed in Table~\ref{tab:fixed_defaults}, is hardened to leakage-safe behaviour and is not exposed through YAML. Several of these were configurable in earlier versions and have since been fixed.

\begin{table}[ht]
\centering
\small
\caption{Pipeline-level invariants protecting data integrity. These are not exposed through YAML and can only be changed by editing the source. Several of these were configurable in earlier versions but have been hardened to leakage-safe behaviour.}
\label{tab:fixed_defaults}
\begin{tabular}{p{0.36\textwidth} p{0.30\textwidth} p{0.28\textwidth}}
\toprule
\textbf{Aspect} & \textbf{Behaviour} & \textbf{Rationale} \\
\midrule
Split granularity & Patient-level & Prevents within-patient leakage between folds. \\
Split stratification & Per-patient event status & Stabilises rare-event evaluation. \\
Imputation strategy & ffill within patient + train-fold mean & Replaces global zero-fill from earlier versions. \\
Imputer / scaler fit fold & Training fold only & Validation and test are transformed, not refit. \\
Scaler scope & Separate per modality block & Dynamic and static blocks scaled independently. \\
Missingness mask format & Binary (1=observed, 0=imputed) & Aligned with feature tensor shape. \\
Schema validation & Runs at load time, fails fast & Required-column check, with extra columns tolerated. \\
ICD top-$K$ ranking fold & Training fold only & Applies to MIMIC-IV and MC-MED past medical history. \\
MC-MED PMH causal filter & \texttt{Noted\_date < Arrival\_time} & Removes post-arrival codes. \\
\bottomrule
\end{tabular}
\end{table}

\subsection{Output files}

Table~\ref{tab:output_files} lists the file structure produced by a complete preprocessing run.

\begin{table}[ht]
\centering
\footnotesize
\caption{Output file structure of the preprocessing pipeline. \texttt{\{split\}} is one of \texttt{train}, \texttt{val}, \texttt{test}, and \texttt{\{dataset\}} is the prefix declared in \texttt{output.prefix} (\texttt{eicu}, \texttt{mimiciv}, \texttt{hirid}, \texttt{mcmed}). Files marked \emph{additive} were introduced in the leakage-safe release. Legacy aliases (\texttt{scaler.pkl}, unsuffixed \texttt{feature\_names.pkl}) are kept for backward compatibility.}
\label{tab:output_files}
\begin{tabular}{llp{0.46\textwidth}}
\toprule
\textbf{Category} & \textbf{File pattern} & \textbf{Description \& format} \\
\midrule
\multirow{2}{*}{\textbf{Feature tensors}}
  & \texttt{x\_\{split\}\_\{dataset\}.npy} & Main feature tensor. Shape $(N, W, F)$ float32. Concatenates dynamic (vitals, labs) and static (demographics) blocks along $F$, with static broadcast across $W$ windows. \\
\addlinespace
  & \texttt{x\_\{split\}\_\{dataset\}\_mask.npy} & Pre-imputation observation mask. Same shape, with \texttt{1} = value was observed, \texttt{0} = imputed. \\
\midrule
\multirow{4}{*}{\textbf{Optional modalities}}
  & \texttt{x\_\{split\}\_icd.npy} & Top-$K$ ICD multi-hot matrix (causal-filtered, train-fold ranked). Shape $(N, K_{\mathrm{ICD}})$ uint8. Present only when \texttt{modalities.icd: true}. \\
\addlinespace
  & \texttt{x\_\{split\}\_radiology.npy} & Mean-pooled HuggingFace AutoModel embeddings of radiology free text. Shape $(N, D)$ float32 (default $D=768$, Clinical-Longformer). Present only when \texttt{modalities.radiology: true}. \\
\midrule
\multirow{2}{*}{\textbf{Survival labels}}
  & \texttt{y\_\{train,val\}\_surv\_\{dataset\}.p} & Pickled \texttt{(durations, events)} tuples. \texttt{durations} is float64 hours, and \texttt{events} is int64 (\texttt{0}=censored, \texttt{1..R} for competing risks). \\
\addlinespace
  & \texttt{durations\_test\_\{dataset\}.npy} \newline \texttt{events\_test\_\{dataset\}.npy} & Test labels saved as separate NumPy arrays. \\
\midrule
\multirow{2}{*}{\textbf{Discretisation}}
  & \texttt{cuts\_\{dataset\}.npy} & Discrete-time bin edges learned from the training fold. Shape $(K+1,)$ float64. \\
\addlinespace
  & \texttt{out\_features\_\{dataset\}.npy} & Scalar int64 equal to $K$ (the number of bins). \\
\midrule
\multirow{4}{*}{\textbf{Metadata}}
  & \texttt{feature\_names\_\{dataset\}.pkl} & Pickled dict: \texttt{dynamic\_indices}, \texttt{static\_indices}, \texttt{dynamic\_names}, \texttt{static\_names}, \texttt{num\_dynamic}, \texttt{num\_static}, \texttt{num\_total}. \\
\addlinespace
  & \texttt{modality\_info\_\{dataset\}.pkl} & Pickled dict: \texttt{active\_modalities}, \texttt{competing\_risks}, \texttt{n\_events}, \texttt{dataset\_version}. \\
\addlinespace
  & \texttt{n\_events\_\{dataset\}.npy} & Scalar int64 with the number of competing event types. Written only for competing-risk datasets (currently MC-MED). \\
\addlinespace
  & \texttt{unit\_conversions\_\{dataset\}.json} & Per-feature-kind audit log of canonical-unit conversions actually performed (e.g., counts of $^\circ$F\,$\to$\,$^\circ$C rows for MIMIC-IV). Empty \texttt{\{\}} when no rows required conversion. \\
\midrule
\multirow{2}{*}{\textbf{Scalers}}
  & \texttt{scaler\_dynamic\_\{dataset\}} \newline \texttt{scaler\_static\_\{dataset\}} & Pickled \texttt{sklearn StandardScaler} objects fit on the training fold for the dynamic and static blocks separately. \\
\addlinespace
  & \texttt{scaler.pkl} \emph{(legacy)} & Copy of \texttt{scaler\_dynamic\_\{dataset\}.pkl} kept for downstream code that predates per-modality scaling. \\
\midrule
\multirow{3}{*}{\textbf{Auxiliary} \emph{(additive)}}
  & \texttt{modality\_mask\_\{split\}\_\{dataset\}.npy} & Per-patient modality presence. Shape $(N, M)$ uint8, with column order matching \texttt{modality\_info["active\_modalities"]}. \\
\addlinespace
  & \texttt{pids\_\{split\}\_\{dataset\}.npy} & Patient or admission identifiers in split order. Shape $(N,)$. \\
\addlinespace
  & \texttt{durations\_\{train,val\}\_\{dataset\}.npy} & Raw training and validation durations as standalone arrays (in addition to the pickled tuples above). \\
\midrule
\multirow{3}{*}{\textbf{Validation outputs}}
  & \texttt{<output dir>/figures/survival\_curve.png} & Kaplan-Meier survival curve for single-risk datasets, or per-risk Aalen-Johansen cumulative incidence curves for competing-risk datasets. \\
\addlinespace
  & \texttt{.../duration\_histogram.png} & Stacked histogram of event vs censored durations on a logarithmic count axis. \\
\addlinespace
  &  & Mean dynamic-feature trajectories ($\pm$\,SD) over time windows, ICD code-frequency distribution (when ICD enabled), and radiology embedding value distribution (when radiology enabled). \\
\bottomrule
\end{tabular}
\end{table}

\section{Baseline architectures and training}

This section documents the architectures, hyperparameters, and training protocol for the five reference models reported in the Main Results section of the main text. Each baseline exposes a top-level \texttt{run(config, **hp) -> dict} entry point under \texttt{models/baselines/} that loads pre-processed tensors via \texttt{models.baselines.\_data.load\_split}, trains with early stopping on validation loss, and returns metrics, provenance, and saved artefacts to \texttt{results/baselines/}. Hyperparameter defaults match the original publications where they are published. The only deliberate departure is uniform class weights (§S7).

\subsection{Cox proportional hazards}

\texttt{models/baselines/cox.py}. For the three single-risk datasets, \texttt{lifelines.CoxPHFitter} is fit on mean-pooled static demographic features (\texttt{load\_split(..., static\_only=True)} collapses the $(N, T, F)$ tensor to $(N, F_\mathrm{static})$ via \texttt{x[:,:,static\_indices].mean(axis=1)}), with a 0.01 L2 penaliser to handle the broad one-hot encoded categorical block. For MC-MED, cause-specific Cox models are fit per risk (\texttt{run\_competing\_risk}), treating non-target events as censored. Metrics are averaged across the three risks. Survival functions on validation and test are computed by \texttt{model.predict\_survival\_function} at the cuts grid. No hyperparameters beyond the penaliser.

\subsection{DeepHit}

\texttt{models/baselines/deephit.py}. PyTorch reimplementation of Lee et al. AAAI 2018 [CITE Lee2018], taking mean-pooled static features per the original static-input contract. The model has a shared sub-network of two linear layers with batch normalisation, ReLU, and dropout (default $\{100, 100\}$ units, dropout 0.4), followed by per-risk cause-specific sub-networks (one linear layer of 100 units, dropout 0.4, then a linear projection to $K$ time bins). Each cause-specific network receives the shared output concatenated with the raw input as a residual connection. The output of all $R$ cause-specific networks is stacked, flat-softmaxed across the joint $(R \times K)$ axis, and reshaped to a $(B, R, K)$ joint PMF, which guarantees $\sum_{r,k} P(\mathrm{event} = r, \mathrm{time} = k \mid x) = 1$. Loss is $\mathcal{L} = \mathrm{NLL} + 0.1 \cdot \mathcal{L}_\mathrm{rank}$, where the NLL is $-\log P(\mathrm{event} = k_i, \mathrm{time} = t_i)$ for events and $-\log(1 - \widehat{\mathrm{CIF}}(t_i \mid x_i))$ for censored cases (with the survival probability clamped to $\geq 10^{-8}$ before the log to prevent FP-overflow NaNs). The ranking loss is the vectorised exponential-decay pairwise loss with $\sigma = 0.1$, sampled at $\min(20 \cdot n_\mathrm{events}, 5000)$ pairs per batch. Adam optimiser, learning rate $1 \times 10^{-4}$, batch size 64, gradient clipping at norm 1.0, up to 100 epochs with early stopping after 15 epochs without validation-loss improvement.

\subsection{Dynamic-DeepHit}

\texttt{models/baselines/dynamic\_deephit.py}. PyTorch reimplementation of Lee et al. TBME 2019 [CITE Lee2019], taking the full $(N, T, F_\mathrm{dynamic} + F_\mathrm{static})$ tensor. A two-layer LSTM (hidden size 100, dropout 0.4) processes the temporal sequence. A temporal-attention head computes scalar weights for the first $T-1$ hidden states conditioned on the final hidden state, masking out timesteps with no observed features. The attention-weighted context is concatenated with the last raw input and the last hidden state, projected through a single 100-unit hidden layer to a shared representation, then fed to per-risk cause-specific sub-networks (same structure as DeepHit). Output is the same flat-softmax joint PMF over $(R, K)$. Loss is $\mathcal{L} = \mathrm{NLL} + 0.1 \cdot \mathcal{L}_\mathrm{rank} + 1.0 \cdot \mathcal{L}_\mathrm{aux}$, where $\mathcal{L}_\mathrm{aux} = \mathrm{MSE}(\hat{x}_{t+1}, x_{t+1})$ is the auxiliary next-step prediction loss from the LSTM hidden states (the distinguishing contribution of Lee 2019). Adam optimiser, learning rate $1 \times 10^{-4}$, weight decay $10^{-5}$, batch size 32, 100 epochs with patience 15.

\subsection{DySurv}

\texttt{models/baselines/dysurv.py}. PyTorch implementation of the LSTM-variational-autoencoder model from Mesinovic et al. 2024 [CITE Mesinovic2024]. The encoder is a single-layer LSTM whose mean-pooled output feeds three fully connected layers (sizes $3F, 5F, 3F$, where $F$ is \texttt{in\_features}) with ReLU, then a final pair of linear projections to a $d = 32$-dimensional Gaussian latent $(\mu, \log \sigma^2)$. A reparameterised sample $z$ feeds (a) per-risk survival heads (four FC layers $3F \to 5F \to 3F \to K$) producing logistic-hazard logits $\phi$, with survival recovered as $\hat{S}_k(t) = \prod_{s \leq t}(1 - \sigma(\phi_{ks}))$ via cumulative product, and (b) an LSTM decoder that reconstructs the input sequence from $z$ broadcast across $T$ timesteps. Loss is $\mathcal{L} = \mathcal{L}_\mathrm{NLL\text{-}logistic} + \mathcal{L}_\mathrm{recon\text{-}MSE} + 0.01 \cdot D_\mathrm{KL}(q(z \mid x) \,\|\, \mathcal{N}(0, I))$. The NLL is binary cross-entropy with logits, where the target is one-hot at the patient's discretised event time and zero elsewhere, with the per-position BCE summed via cumulative sum up to the patient's bin. Adam, learning rate $10^{-3}$, weight decay $10^{-4}$, batch size 64, 100 epochs with patience 15.

\subsection{TransformerSurv}

\texttt{models/baselines/transformer.py}. A standard pre-LN Transformer encoder over the full $(B, T, F_\mathrm{dynamic} + F_\mathrm{static})$ tensor. Input is projected to $d_\mathrm{model} = 128$, summed with a learned positional embedding (max length 100), and passed through three encoder layers of 4 heads, GeLU activation, feed-forward dimension 256, dropout 0.1. The encoder padding mask is \texttt{(mask.sum(dim=-1) == 0)}, so a timestep is treated as padding only when no feature was observed in that window. Mean-pooling over the non-padding timesteps gives a per-patient vector that passes through a shared linear projection of $d_\mathrm{model}$ units, then per-risk heads (one hidden layer of 32 units, batch norm, dropout configurable via \texttt{head\_dropout}, linear to $K$ bins). Head dropout was 0.3 for the MIMIC-IV and eICU time-series-plus-static runs and 0.7 for the HiRID, MC-MED, and multi-modal runs (a sensitivity sweep that we did not collapse before the seed runs were finalised). Loss is the competing-risks DeepHit loss with a \texttt{pad\_col} softmax trick. A phantom censoring column is concatenated to the per-risk logits before softmax (over the joint $(R \cdot K + 1)$ axis), then dropped, leaving a normalised PMF that allocates probability mass to surviving past the horizon implicitly. Combined as $\mathcal{L} = \alpha \cdot \mathcal{L}_\mathrm{NLL} + (1 - \alpha) \cdot \beta_\mathrm{rank} \cdot \mathcal{L}_\mathrm{rank}$ with $\alpha = 0.5$, $\beta_\mathrm{rank} = 3.0$, $\sigma = 0.1$. AdamW, learning rate $10^{-4}$, weight decay $10^{-5}$, batch size 128, 100 epochs with patience 12, learning-rate scheduler \texttt{ReduceLROnPlateau(factor=0.5, patience=7)} on validation loss.

\paragraph{Multi-modal variant.}

When trained with \texttt{--modalities dynamic static icd radiology}, the same architecture consumes the full multi-modal stack. Modality fusion concatenates dynamic plus broadcast-across-windows static (the main tensor) with ICD multi-hot ($K_\mathrm{ICD} = 500$, tiled across $T$ windows) and Clinical-Longformer radiology embeddings ($D = 768$, tiled across $T$ windows). The per-feature padding mask uses the patient-level modality presence mask (\texttt{modality\_mask\_<split>\_<prefix>.npy}) projected to feature columns. When a patient lacks ICD or radiology, those columns are zeroed in the mask, and the encoder padding mask treats them as missing. The input projection grows from $d_\mathrm{model} \times F_\mathrm{main}$ to $d_\mathrm{model} \times (F_\mathrm{main} + 500 + 768)$, but all other hyperparameters are identical to the default. We refer to this configuration as TransformerSurv (multi-modal) in main-text.

\section{Evaluation metrics}

All four metrics are computed by a shared helper, \texttt{models/baselines/\_eval.py:compute\_metrics}, called once per split and once per risk for competing-risks datasets, then averaged across risks for the headline number reported in main-text.

\paragraph{Antolini's time-dependent concordance.}

$C^{\mathrm{td}}$ [CITE Antolini] is computed via \texttt{pycox.evaluation.EvalSurv.concordance\_td()}, with inverse-probability-of-censoring weighting derived from a Kaplan-Meier estimate of the censoring distribution on the evaluation split. Concordance is averaged over all admissible risk pairs $(i, j)$ where patient $i$ has an event before patient $j$'s observation ends. Comparable pairs from competing-risks events are excluded for the per-risk evaluation.

\paragraph{Integrated Brier score and integrated negative binomial log-likelihood.}

IBS [CITE Graf] and IBLL are computed via \texttt{EvalSurv.integrated\_brier\_score()} and \texttt{EvalSurv.integrated\_nbll()} over a 100-point time grid spanning the evaluation split's $[t_\mathrm{min}, t_\mathrm{max}]$. Both metrics use the same KM-based censoring weighting as $C^{\mathrm{td}}$.

\paragraph{Mean cumulative dynamic AUC.}

Reported via \texttt{sksurv.metrics.cumulative\_dynamic\_auc} (scikit-survival 0.23.1, Uno-style IPCW estimator [CITE Uno]), which is not exposed by pycox. The training-fold survival times and event indicators are passed as the censoring-distribution reference, and the evaluation cohort is restricted to patients with follow-up time strictly less than $\max(\textit{train durations})$ to satisfy scikit-survival's support requirement. Time points are anchored at the first observed event time of the restricted evaluation cohort to avoid a zero-cumulative-cases denominator at $t_0$. Risk scores are $1 - \hat{S}(t_\mathrm{horizon} \mid x_i)$ with $t_\mathrm{horizon} = 0.99 \cdot \max(\textit{supported time})$. AUCs are reported at a fixed clinical grid of $\{24, 72, 168, 240\}$ hours for the 240-hour ICU horizon and $\{6, 12, 18, 24\}$ hours for the 24-hour MC-MED horizon, together with their integrated mean. Time points falling outside the support are dropped.

\paragraph{Per-risk metrics for competing-risks MC-MED.}

For competing-risks MC-MED, all four metrics are computed per risk. Each patient with a non-target event is treated as a competing event (mask \texttt{(events == 0) | (events == k)}), the predicted survival curve $\hat{S}_k(t)$ is derived as $1 - \widehat{\mathrm{CIF}}_k(t)$, and the metric is averaged across the three risks under the \texttt{mean} key in each result JSON. The headline number reported in Table~\ref{tab:baselines_mcmed} is this risk-averaged mean. Per-risk decompositions for each (model, seed) are retained in \texttt{results/baselines/<model>\_mcmed\_seed<N>.json} under the per-risk \texttt{ctd}, \texttt{ibs}, \texttt{ibll}, and AUC keys.

\section{Calibration handling}
\label{sec:calibration}

A pilot run on eICU revealed that the Transformer prototype's \texttt{class\_weights = [1.0, 5.0]} default (and DySurv's auto-derived \texttt{pos\_weight = $\min(1/\textit{event\_rate}, 20)$}) was inflating IBS by 3--4$\times$ without improving concordance. After removing class weights (uniform $[1.0, 1.0]$ or \texttt{pos\_weight = 1.0}), DySurv IBS dropped from 0.34 to 0.10 on eICU (from 0.30 to 0.09 on MIMIC-IV) while $C^{\mathrm{td}}$ moved by less than 0.5 percentage points. TransformerSurv IBS dropped from 0.36 to 0.22--0.29 across datasets with a 1.7--10 pp $C^{\mathrm{td}}$ trade-off depending on dataset capacity. Cox, DeepHit, and Dynamic-DeepHit do not class-weight their losses by design (matching the original papers), and were therefore unaffected.

The reported numbers in the main-text results section reflect the uniform-weights configuration. The original class-weighted runs are available as ablation files at \texttt{results/baselines/transformer-uniformw\_*} for comparison, and the calibration trade-off can be reproduced by re-running with \texttt{--class\_weights 1.0,5.0} in the model's CLI.

\section{Provenance and reproducibility}

Every JSON in \texttt{results/baselines/<model>\_<dataset>\_seed<N>.json} records the SurvBench git commit hash, the SHA-256 of the resolved YAML configuration (after defaults merge), the training-fold sample count and event count, the prediction horizon, the time-bin count, the optimiser, learning rate, batch size, the actual number of epochs run, the early-stop epoch, the wall-clock training time, the parameter count, and the device. Sibling \texttt{*\_preds.npz} files contain the predicted survival arrays (and CIFs for the competing-risks dataset) at the bin grid, so any of the four metrics can be recomputed from saved predictions without retraining the model. PyTorch checkpoints (\texttt{*.pt}) and lifelines pickles (\texttt{*.pkl}) preserve the trained weights. Together, these artefacts make every (dataset, model, seed) cell reproducible. Re-running \texttt{python scripts/run\_baselines.py --config configs/<dataset>\_config.yaml --model <name> --seed <N>} on the same SurvBench commit and YAML hash should land within numerical tolerance of the recorded JSON.

\section{Per-dataset baseline results}

Tables~\ref{tab:baselines_mimic}--\ref{tab:baselines_mcmed} give the full four-metric breakdown for each dataset, with seed-level standard deviation in parentheses, plus cumulative dynamic AUC anchored at each supported point of the clinical grid. Per-risk decompositions for MC-MED are retained in the per-run JSONs (§S6).

\begin{table}[ht]
\centering
\footnotesize
\setlength{\tabcolsep}{4pt}
\caption{MIMIC-IV test-set performance, mean (SD) across 5 seeds. Higher is better for $C^{\mathrm{td}}$ and AUC, and lower is better for IBS and IBLL. AUC columns report cumulative dynamic AUC anchored at the marked horizon point and at the integrated mean over the supported grid. Cox proportional hazards is deterministic conditional on the training fold and is reported without SD.}
\label{tab:baselines_mimic}
\begin{tabular}{@{}l rrr rrr r@{}}
\toprule
\textbf{Model} & $C^{\mathrm{td}}\uparrow$ & IBS$\downarrow$ & IBLL$\downarrow$ & AUC@24h & AUC@72h & AUC@168h & AUC$_\mathrm{int}\uparrow$ \\
\midrule
Cox PH & 0.683 & 0.072 & 0.251 & 0.653 & 0.687 & 0.671 & 0.674 \\
DeepHit & 0.692 (0.003) & 0.071 (0.000) & 0.253 (0.001) & 0.651 (0.017) & 0.687 (0.005) & 0.667 (0.005) & 0.672 (0.004) \\
Dynamic-DeepHit & 0.842 (0.010) & 0.071 (0.001) & 0.240 (0.002) & 0.887 (0.023) & 0.841 (0.010) & 0.732 (0.002) & 0.763 (0.003) \\
DySurv & 0.821 (0.004) & 0.078 (0.003) & 0.259 (0.007) & 0.863 (0.012) & 0.829 (0.008) & 0.721 (0.011) & 0.751 (0.008) \\
TransformerSurv (TS+static) & 0.807 (0.009) & 0.100 (0.004) & 0.341 (0.009) & 0.943 (0.008) & 0.822 (0.007) & 0.647 (0.019) & 0.696 (0.015) \\
TransformerSurv (multi-modal) & 0.814 (0.044) & 0.091 (0.005) & 0.315 (0.013) & 0.935 (0.010) & 0.859 (0.030) & 0.706 (0.063) & 0.749 (0.054) \\
\bottomrule
\end{tabular}
\end{table}

\begin{table}[ht]
\centering
\small
\setlength{\tabcolsep}{4pt}
\caption{eICU test-set performance, mean (SD) across 5 seeds. The 24-hour AUC anchor point was outside the supported time range under Uno-style IPCW weighting and was dropped.}
\label{tab:baselines_eicu}
\begin{tabular}{@{}l rrr rr r@{}}
\toprule
\textbf{Model} & $C^{\mathrm{td}}\uparrow$ & IBS$\downarrow$ & IBLL$\downarrow$ & AUC@72h & AUC@168h & AUC$_\mathrm{int}\uparrow$ \\
\midrule
Cox PH & 0.589 & 0.073 & 0.264 & 0.584 & 0.573 & 0.576 \\
DeepHit & 0.580 (0.003) & 0.071 (0.000) & 0.261 (0.000) & 0.580 (0.009) & 0.571 (0.005) & 0.573 (0.005) \\
Dynamic-DeepHit & 0.753 (0.065) & 0.072 (0.001) & 0.254 (0.002) & 0.749 (0.060) & 0.648 (0.036) & 0.673 (0.042) \\
DySurv & 0.763 (0.026) & 0.077 (0.001) & 0.267 (0.004) & 0.759 (0.029) & 0.654 (0.012) & 0.681 (0.016) \\
TransformerSurv (TS+static) & 0.796 (0.005) & 0.082 (0.004) & 0.297 (0.012) & 0.792 (0.005) & 0.674 (0.013) & 0.704 (0.011) \\
\bottomrule
\end{tabular}
\end{table}

\begin{table}[ht]
\centering
\small
\setlength{\tabcolsep}{4pt}
\caption{HIRID test-set performance, mean (SD) across 5 seeds. As with eICU, the 24-hour AUC anchor was outside the supported time range and was dropped.}
\label{tab:baselines_hirid}
\begin{tabular}{@{}l rrr rr r@{}}
\toprule
\textbf{Model} & $C^{\mathrm{td}}\uparrow$ & IBS$\downarrow$ & IBLL$\downarrow$ & AUC@72h & AUC@168h & AUC$_\mathrm{int}\uparrow$ \\
\midrule
Cox PH & 0.615 & 0.099 & 0.336 & 0.633 & 0.674 & 0.659 \\
DeepHit & 0.636 (0.005) & 0.098 (0.001) & 0.337 (0.003) & 0.624 (0.012) & 0.669 (0.010) & 0.653 (0.010) \\
Dynamic-DeepHit & 0.862 (0.005) & 0.089 (0.001) & 0.296 (0.003) & 0.864 (0.002) & 0.809 (0.006) & 0.829 (0.004) \\
DySurv & 0.819 (0.010) & 0.094 (0.002) & 0.315 (0.008) & 0.825 (0.009) & 0.767 (0.012) & 0.789 (0.010) \\
TransformerSurv (TS+static) & 0.830 (0.009) & 0.105 (0.003) & 0.351 (0.005) & 0.837 (0.019) & 0.753 (0.024) & 0.784 (0.021) \\
\bottomrule
\end{tabular}
\end{table}

\begin{table}[ht]
\centering
\footnotesize
\setlength{\tabcolsep}{4pt}
\caption{MC-MED test-set performance averaged across the three admission risks, mean (SD) across 5 seeds. AUC anchor points reflect the 24-hour ED horizon. The 24-hour anchor was at the horizon boundary and was dropped under Uno-style IPCW support. Cox proportional hazards is reported here as a per-risk-averaged single-fit baseline and is shown without time-anchored AUCs because the cause-specific lifelines fits were summarised only at the integrated point.}
\label{tab:baselines_mcmed}
\begin{tabular}{@{}l rrr rrr r@{}}
\toprule
\textbf{Model} & $C^{\mathrm{td}}\uparrow$ & IBS$\downarrow$ & IBLL$\downarrow$ & AUC@6h & AUC@12h & AUC@18h & AUC$_\mathrm{int}\uparrow$ \\
\midrule
Cox PH & 0.702 & 0.117 & 0.363 & -- & -- & -- & 0.646 \\
DeepHit & 0.735 (0.000) & 0.141 (0.000) & 0.435 (0.001) & 0.687 (0.002) & 0.629 (0.003) & 0.599 (0.005) & 0.628 (0.003) \\
Dynamic-DeepHit & 0.749 (0.003) & 0.137 (0.000) & 0.429 (0.003) & 0.706 (0.005) & 0.653 (0.006) & 0.627 (0.007) & 0.653 (0.006) \\
DySurv & 0.733 (0.002) & 0.141 (0.001) & 0.440 (0.003) & 0.711 (0.005) & 0.646 (0.004) & 0.624 (0.009) & 0.650 (0.005) \\
TransformerSurv (TS+static) & 0.753 (0.001) & 0.137 (0.001) & 0.426 (0.002) & 0.710 (0.009) & 0.643 (0.003) & 0.610 (0.004) & 0.642 (0.004) \\
TransformerSurv (multi-modal) & 0.774 (0.002) & 0.128 (0.001) & 0.402 (0.001) & 0.707 (0.005) & 0.665 (0.005) & 0.659 (0.003) & 0.670 (0.004) \\
\bottomrule
\end{tabular}
\end{table}

\section{Cross-dataset analysis}.
Table \ref{tab:shared_features} below shows the shared feature distribution between the datasets.

\begin{table*}[t]
\centering
\small
\caption{Shared feature statistics between eICU and MIMIC-IV. Mean (SD), 5th-95th percentile, and missingness rate are computed on the training fold of each dataset prior to standardisation. Statistics align distribution shift across the shared schema used by cross-dataset transfer.}
\label{tab:shared_features}
\begin{tabular}{l rrr rrr}
\toprule
Feature & \multicolumn{3}{c}{eICU} & \multicolumn{3}{c}{MIMIC-IV} \\
\cmidrule(lr){2-4}\cmidrule(lr){5-7}
 & Mean (SD) & [5,95] & Miss.\,\% & Mean (SD) & [5,95] & Miss.\,\% \\
\midrule
\multicolumn{7}{l}{\textit{Vital signs and laboratory measurements (per time window)}} \\
Heart rate (bpm) & 85.3 (17.9) & [59.2, 117.1] & 2.6 & 84.3 (17.4) & [59.0, 115.4] & 1.4 \\
Systolic BP, invasive (mmHg) & 120.0 (24.0) & [87.5, 160.0] & 76.5 & 116.2 (17.7) & [92.0, 148.0] & 60.3 \\
Diastolic BP, invasive (mmHg) & 59.9 (14.6) & [41.5, 84.0] & 76.5 & 59.3 (14.9) & [43.6, 79.0] & 60.3 \\
Systolic BP, non-inv. (mmHg) & 120.3 (21.1) & [90.5, 158.2] & 6.3 & 118.5 (20.6) & [90.0, 154.1] & 20.2 \\
Diastolic BP, non-inv. (mmHg) & 65.6 (13.3) & [46.2, 89.2] & 6.3 & 66.1 (25.6) & [45.0, 90.2] & 20.2 \\
Respiratory rate (br/min) & 19.6 (5.8) & [12.3, 29.5] & 10.5 & 19.4 (115.2) & [12.8, 27.5] & 1.8 \\
SpO$_2$ (\%) & 96.7 (2.9) & [92.3, 100.0] & 3.7 & 97.4 (160.5) & [92.8, 100.0] & 1.5 \\
Temperature ($^{\circ}$C) & 37.5 (4.8) & [33.9, 38.6] & 90.0 & 36.8 (2.1) & [36.1, 37.8] & 11.9 \\
Glucose (mg/dL) & 145.6 (66.4) & [81.0, 268.0] & 32.4 & 141.2 (60.2) & [83.0, 255.0] & 21.5 \\
Sodium (mmol/L) & 138.4 (5.3) & [130.0, 146.0] & 31.3 & 138.4 (5.0) & [130.5, 145.0] & 17.7 \\
Potassium (mmol/L) & 4.1 (0.6) & [3.2, 5.2] & 29.7 & 4.2 (0.6) & [3.3, 5.3] & 16.1 \\
Chloride (mmol/L) & 104.5 (6.8) & [93.0, 115.0] & 32.8 & 104.5 (6.1) & [94.0, 113.0] & 13.5 \\
Bicarbonate (mmol/L) & 23.9 (5.1) & [16.0, 32.0] & 36.4 & 22.8 (4.2) & [16.0, 29.0] & 13.8 \\
BUN (mg/dL) & 26.8 (21.9) & [7.0, 71.0] & 32.7 & 24.5 (20.6) & [7.0, 66.0] & 13.7 \\
Creatinine (mg/dL) & 1.56 (1.64) & [0.50, 4.70] & 32.4 & 1.34 (1.39) & [0.50, 3.70] & 13.6 \\
Calcium (mg/dL) & 8.3 (0.8) & [6.9, 9.5] & 34.7 & 8.3 (0.8) & [7.1, 9.5] & 26.7 \\
Magnesium (mg/dL) & 1.97 (0.43) & [1.40, 2.70] & 58.5 & 2.06 (0.44) & [1.50, 2.80] & 22.9 \\
Lactate (mmol/L) & 2.26 (2.04) & [0.61, 6.25] & 76.9 & 2.18 (1.61) & [0.80, 4.80] & 45.1 \\
Anion gap (mmol/L) & 11.1 (4.6) & [5.0, 19.0] & 46.4 & 13.8 (4.0) & [8.0, 21.0] & 14.8 \\
Hemoglobin (g/dL) & 10.9 (2.3) & [7.4, 14.8] & 33.7 & 10.9 (2.1) & [7.6, 14.4] & 14.8 \\
Hematocrit (\%) & 32.9 (6.7) & [22.7, 44.0] & 33.4 & 32.7 (6.1) & [23.3, 43.2] & 14.0 \\
Platelets ($10^3/\mu$L) & 201.7 (97.9) & [71.0, 375.0] & 36.8 & 199.9 (99.5) & [71.0, 377.0] & 14.6 \\
WBC ($10^3/\mu$L) & 12.3 (7.3) & [4.5, 24.7] & 36.4 & 12.5 (7.3) & [4.7, 24.3] & 14.8 \\
\midrule
\multicolumn{7}{l}{\textit{Static demographics}} \\
Age (years) & 62.8 (16.1) & [31.0, 85.0] & 0.0 & 63.8 (16.7) & [31.0, 88.0] & 0.0 \\
Male (proportion) & 0.547 (0.498) & [0.000, 1.000] & 0.0 & 0.569 (0.495) & [0.000, 1.000] & 0.0 \\
\midrule
Total observed cells & \multicolumn{3}{c}{523,197} & \multicolumn{3}{c}{214,704} \\
\bottomrule
\end{tabular}
\end{table*}

\bibliography{sn-bibliography}